\documentclass[journal]{IEEEtran}

\usepackage[utf8]{inputenc}
\usepackage[english]{babel}
\usepackage{amsthm} 

\usepackage{algorithm}
\usepackage{algpseudocode}

\usepackage{amssymb}  
\usepackage{amsmath}   
\usepackage{graphicx}  
\usepackage{subcaption}  
\usepackage{multirow}  
\usepackage{xcolor}  

\def\b{\ensuremath\boldsymbol}

\usepackage{url}
\ifCLASSINFOpdf
\else
\fi
\begin{document}
%
\title{Roweis Discriminant Analysis: \\A Generalized Subspace Learning Method}
%
%
%

\author{Benyamin Ghojogh,
        Fakhri Karray,
        Mark Crowley
\thanks{All Authors are with the department of electrical and computer engineering, University of Waterloo, Waterloo, ON, Canada.}}

%
%

\markboth{}%
{Shell \MakeLowercase{\textit{et al.}}: Bare Demo of IEEEtran.cls for IEEE Journals}
%



\maketitle

\begin{abstract}
We present a new method which generalizes subspace learning based on eigenvalue and generalized eigenvalue problems. This method, Roweis Discriminant Analysis (RDA), is named after Sam Roweis to whom the field of subspace learning owes significantly. RDA is a family of infinite number of algorithms where Principal Component Analysis (PCA), Supervised PCA (SPCA), and Fisher Discriminant Analysis (FDA) are special cases. One of the extreme special cases, which we name Double Supervised Discriminant Analysis (DSDA), uses the labels twice; it is novel and has not appeared elsewhere. We propose a dual for RDA for some special cases. We also propose kernel RDA, generalizing kernel PCA, kernel SPCA, and kernel FDA, using both dual RDA and representation theory. Our theoretical analysis explains previously known facts such as why SPCA can use regression but FDA cannot, why PCA and SPCA have duals but FDA does not, why kernel PCA and kernel SPCA use kernel trick but kernel FDA does not, and why PCA is the best linear method for reconstruction. Roweisfaces and kernel Roweisfaces are also proposed generalizing eigenfaces, Fisherfaces, supervised eigenfaces, and their kernel variants. We also report experiments showing the effectiveness of RDA and kernel RDA on some benchmark datasets.
\end{abstract}


%
\IEEEpeerreviewmaketitle

\section{Introduction}
%
%
%
%

\IEEEPARstart{S}{ubspace} and manifold learning, also referred to as representation learning \cite{bengio2013representation,zhong2016overview}, are very useful in machine learning and pattern analysis for feature extraction, data visualization, and dimensionality reduction \cite{ghojogh2019feature}.
The subspace of data is a lower-dimensional space than the input space of data which can appropriately represent the data with the smallest possible representation error. In other words, the data usually exist on a submanifold or subspace with a lower intrinsic dimensionality \cite{wang2012geometric,warner2013foundations}.

The submanifold of data can be either linear or nonlinear. 
Different linear and nonlinear methods have been proposed in the literature for subspace and manifold learning. 
Principal Component Analysis (PCA) \cite{jolliffe2011principal}, first proposed in \cite{pearson1901liii}, was one of the first methods in linear subspace learning. 
The Fisher Discriminant Analysis (FDA) \cite{friedman2001elements}, first proposed in \cite{fisher1936use}, was one of the first linear supervised subspace learning methods. Both PCA and FDA were based on the scatter of data. 
Metric Multi-Dimensional Scaling (MDS) \cite{cox2000multidimensional} was a linear method which tried to preserve the similarity of the data points. In later approaches after MDS, the cost function in MDS was changed to preserving the distances of points \cite{lee2007nonlinear}, yielding to Sammon mapping \cite{sammon1969nonlinear}.
Sammon mapping can probably be considered as the first nonlinear subspace learning method.  
Another approach to handle nonlinearity of data is to modify the data instead of changing the linear algorithm. That was the perspective of kernel PCA \cite{scholkopf1997kernel,scholkopf1998nonlinear} which used the dual of PCA and the kernel trick \cite{hofmann2008kernel} to pull the data to the feature space hoping that it becomes roughly linear in that space. Kernel FDA \cite{mika1999fisher,mika2000invariant} was also proposed for handling the nonlinear data in a supervised manner. However, it did not use the kernel trick (which we will explain why in this paper) but used representation theory \cite{alperin1993local} instead. 
Recently, deep FDA \cite{diaz2017deep,diaz2019deep} was proposed which uses a least squares approach \cite{ye2007least,zhang2010regularized}.

After Sammon mapping, there was not any important method, which is actually nonlinear without changing the data like kernel PCA/FDA, until Isomap \cite{tenenbaum2000global} and Locally Linear Embedding (LLE) \cite{roweis2000nonlinear}. 
The former used geodesic distance rather than Euclidean distance in the kernel of MDS \cite{ham2004kernel,strange2014open} and the latter reconstructed every point by its $k$-Nearest Neighbors ($k$-NN) to locally fit the data \cite{saul2003think}. 
Stochastic Neighbor Embedding (SNE) \cite{hinton2003stochastic} was a probabilistic approach to manifold learning where the probability of a point being neighbor of others was tried to be preserved. While the Gaussian distribution was used in SNE, the Student-$t$ distribution was used in the embedded space in $t$-SNE method in order to tackle the crowding problem \cite{maaten2008visualizing}. 
Gradually, deep learning became popular where the Deep Belief Network (DBN) \cite{hinton2006reducing} was proposed in order to learn the latent subspace of data in an undercomplete autoencoder. 
One of the most recent important subspace learning methods was Supervised PCA (SPCA) \cite{barshan2011supervised}. The SPCA made use of the empirical estimation of the Hilbert-Schmidt Independence Criterion (HSIC) \cite{gretton2005measuring}. The HSIC computes the dependence of two random variables by calculating their correlation in the feature space. The SPCA tried to maximize the dependence of projected data and the labels in order to use the information of labels for better embedding. 
Recently, supervised random projection \cite{karimi2018srp} used low-rank kernel approximation in the formulation of SPCA for better efficiency.

In this paper, we propose a generalized subspace learning method, named Roweis Discriminant Analysis (RDA), named after Sam Roweis (1972--2010) who contributed significantly to subspace and manifold learning. He proposed many important methods in subspace/manifold learning including LLE \cite{roweis2000nonlinear,saul2003think}, SNE \cite{hinton2003stochastic}, and metric learning by class collapsing \cite{globerson2006metric}. 
The proposed RDA is a family of infinite number of subspace learning methods including PCA, FDA, and SPCA which have been proposed in the literature. It is a generalized method and contains many methods which use linear projection into a lower dimensional subspace. 
The main contributions of our paper can be listed as the following: 
\begin{enumerate}
\item proposing RDA as a generalized subspace learning method based on eigenvalue and generalized eigenvalue problems,
\item proposing Double Supervised Discriminant Analysis (DSDA), which is not yet proposed in the literature to the best of our knowledge, as one of the extreme cases in RDA,
\item proposing RDA in the feature space to have kernel RDA which generalizes kernel PCA, kernel SPCA, and kernel FDA,
\item explaining the reasons behind some of the characteristics of PCA, SPCA, and FDA such as (I) why SPCA can be used for both classification and regression but FDA is only for classification, (II) why PCA and SPCA have their dual methods but FDA does not have a dual, (III) why kernel PCA and kernel SPCA can use kernel trick but kernel FDA uses representation theory rather than kernel trick, (IV) why PCA is the best linear method for reconstruction,
\item proposing Roweisfaces and kernel Roweisfaces for demonstrating the generalizability of the RDA approach for the eigenfaces \cite{turk1991eigenfaces,turk1991face}, kernel eigenfaces \cite{yang2000face}, Fisherfaces \cite{belhumeur1997eigenfaces}, kernel Fisherfaces \cite{yang2002kernel} and supervised eigenfaces \cite{barshan2011supervised,ghojogh2019unsupervised}.
\end{enumerate}

The remainder of the paper is organized as follows. Section \ref{section_subspace} introduces projection into a subspace and the reconstruction of data after projection. Two general forms of optimization for subspace learning are also introduced. In Section \ref{section_PCA_FDA_SPCA}, the theory of PCA, FDA, and SPCA is briefly reviewed. The proposed RDA is explained in Section \ref{section_RDA}. A dual is also proposed for the RDA in Section \ref{section_dual_RDA} but for some special cases.
Section \ref{section_kernel_RDA} explains two types of kernel RDA, one based on kernel trick and another using representation theory. The experimental results are reported in Section \ref{section_experiments} where the Roweisfaces are also proposed. Section \ref{section_conclusion} summarizes and concludes the article. 


\section{Subspace and Projection}\label{section_subspace}

Let $n$, $n_t$, $d$, $c$, $n_j$, $\b{x}_i$, $\b{x}_i^{(j)}$, and $\b{x}_{t,i}$ denote the training sample size, test (out-of-sample) sample size, dimensionality of data, number of classes, sample size of the $j$-th class, the $i$-th training instance, the $i$-th training instance in the $j$-th class, and the $i$-th test instance, respectively. 
We stack the $n$ training data points column-wise in a matrix $\b{X} = [\b{x}_1, \dots, \b{x}_n] \in \mathbb{R}^{d \times n}$ and similarly for the $n_t$ out-of-sample data points as $\b{X}_t = [\b{x}_{t,1}, \dots, \b{x}_{t,n_t}] \in \mathbb{R}^{d \times n_t}$.

Let $\b{U} \in \mathbb{R}^{d \times d}$ be the projection matrix whose columns $\{\b{u}_j\}_{j=1}^d$ are the projection directions spanning the desired subspace so the subspace is the column-space of $\b{U}$. If we truncate the projection matrix to have $\mathbb{R}^{d \times p} \ni \b{U} = [\b{u}_1, \dots, \b{u}_p]$, the subspace is spanned by $p$ projection directions and it will be a $p$ dimensional subspace where $p \leq d$. 
We want the projection directions to be orthonormal to capture different information; therefore:
\begin{align}
\b{U}^\top \b{U} = \b{I},
\end{align}
where $\b{I}$ is the identity matrix.

The projection of training data into the subspace and its reconstruction after the projection are \cite{wang2012geometric}:
\begin{align}
&\mathbb{R}^{p \times n} \ni \widetilde{\b{X}} := \b{U}^\top \b{X}, \label{equation_projection_severalPoint} \\
&\mathbb{R}^{d \times n} \ni \widehat{\b{X}} := \b{U}\b{U}^\top \b{X} = \b{U}\widetilde{\b{X}}, \label{equation_reconstruction_severalPoint}
\end{align}
respectively, where $\widetilde{\b{X}} = [\widetilde{\b{x}}_1, \dots, \widetilde{\b{x}}_n]$ and $\widehat{\b{X}} = [\widehat{\b{x}}_1, \dots, \widehat{\b{x}}_n]$.
The projection and reconstruction of the out-of-sample data are:
\begin{align}
&\mathbb{R}^{p \times n_t} \ni \widetilde{\b{X}}_t := \b{U}^\top \b{X}_t, \label{equation_outOfSample_projection_severalPoint} \\
&\mathbb{R}^{d \times n} \ni \widehat{\b{X}}_t := \b{U}\b{U}^\top \b{X}_t = \b{U}\widetilde{\b{X}}_t, \label{equation_outOfSample_reconstruction_severalPoint}
\end{align}
respectively, where $\widetilde{\b{X}}_t = [\widetilde{\b{x}}_{t,1}, \dots, \widetilde{\b{x}}_{t,n_t}]$ and $\widehat{\b{X}}_t = [\widehat{\b{x}}_{t,1}, \dots, \widehat{\b{x}}_{t,n_t}]$.
Note that in subspace learning, it is recommended to center the data, either training or out-of-sample, using the training mean. If we do that, the training mean should be added back to the reconstructed data.
If the mean of training data is:
\begin{align}\label{equation_total_mean}
\b{\mu} := \frac{1}{n} \sum_{i=1}^n \b{x}_i,
\end{align}
the centered training data are:
\begin{align}\label{equation_centered_training_data}
\mathbb{R}^{d \times n} \ni \breve{\b{X}} := \b{X} \b{H} = \b{X} - \b{\mu},
\end{align}
where $\breve{\b{X}} = [\breve{\b{x}}_1, \dots, \breve{\b{x}}_n] = [\b{x}_1 - \b{\mu}, \dots, \b{x}_n - \b{\mu}]$ and: 
\begin{align}
\mathbb{R}^{n \times n} \ni \b{H} := \b{I} - (1/n) \b{1}\b{1}^\top,
\end{align}
is the centering matrix.
The covariance matrix, or the total scatter, is defined as:
\begin{equation}\label{equation_total_scatter}
\begin{aligned}
\mathbb{R}^{n \times n} \ni \b{S}_T &:= \sum_{i=1}^n (\b{x}_i - \b{\mu})(\b{x}_i - \b{\mu})^\top \\
&= \breve{\b{X}} \breve{\b{X}}^\top \overset{(\ref{equation_centered_training_data})}{=} \b{X} \b{H} \b{H} \b{X}^\top = \b{X} \b{H} \b{X}^\top, 
\end{aligned}
\end{equation}
where it is noticed that the centering matrix is symmetric and idempotent. 

If we project the centered training data, we have:
\begin{align}\label{equation_squared_length_reconstruction}
||\widehat{\b{X}}||_F^2 = ||\b{U}\b{U}^\top \breve{\b{X}}||_F^2 = \textbf{tr}(\b{U}^\top \b{S}_T\, \b{U}),
\end{align}
where $||.||_F$ and $\textbf{tr}(.)$ denote the Frobenius norm and trace of matrix, respectively.
Hence, $\textbf{tr}(\b{U}^\top \b{S}\, \b{U})$ is the squared length of the reconstruction.
This term can also be interpreted as the variance of the projected data according to quadratic characteristic of variance. 
We want the variance of projection to be maximized where the projection matrix is orthogonal; otherwise, the optimization problem is ill-defined. If we consider a general scatter of data, $\b{S} \in \mathbb{R}^{d \times d}$, the optimization is:
\begin{equation}\label{equation_optimization_generalForm_eigenvalue}
\begin{aligned}
& \underset{\b{U}}{\text{maximize}}
& & \textbf{tr}(\b{U}^\top \b{S}\, \b{U}), \\
& \text{subject to}
& & \b{U}^\top \b{U} = \b{I},
\end{aligned}
\end{equation}
where the constraint makes the problem well-defined.
The Lagrangian \cite{boyd2004convex} of the problem is:
\begin{align*}
\mathcal{L} = \textbf{tr}(\b{U}^\top \b{S}\, \b{U}) - \textbf{tr}\big(\b{\Lambda}^\top (\b{U}^\top \b{U} - \b{I})\big),
\end{align*}
where $\b{\Lambda} \in \mathbb{R}^{p \times p}$ is a diagonal matrix $\textbf{diag}([\lambda_1, \dots, \lambda_p]^\top)$ including the Lagrange multipliers. 
Setting the derivative of Lagrangian to zero gives:
\begin{align}
& \mathbb{R}^{d \times p} \ni \frac{\partial \mathcal{L}}{\partial \b{U}} = 2\b{S} \b{U} - 2\b{U} \b{\Lambda} \overset{\text{set}}{=} \b{0}  \implies \b{S} \b{U} = \b{U} \b{\Lambda}, \label{equation_scatter_eigendecomposition}
\end{align}
which is the eigenvalue problem for $\b{S}$ where the columns of $\b{U}$ and the diagonal of $\b{\Lambda}$ are the eigenvectors and eigenvalues of $\b{S}$, respectively \cite{ghojogh2019eigenvalue}. The eigenvectors and eigenvalues are sorted from the leading (largest eigenvalue) to the trailing (smallest eigenvalue) because we are maximizing in the optimization problem (the reason lies in the second order condition).

We can also have two types of scatters, e.g., $\b{S}_1 \in \mathbb{R}^{n \times n}$ and $\b{S}_2 \in \mathbb{R}^{n \times n}$. 
As the scatter matrix is symmetric and positive semi-definite, we can decompose it as:
\begin{align}
\b{S}_2 \overset{(a)}{=} \b{\Psi}_S\, \b{\Omega}_S \b{\Psi}_S^\top = \b{\Psi}_S\, \b{\Omega}_S^{(1/2)} \b{\Omega}_S^{(1/2)} \b{\Psi}_S^\top \overset{(b)}{=} \b{\Delta}^\top \b{\Delta},
\end{align}
where $(a)$ is because of Singular Value Decomposition (SVD) and $(b)$ is for $\mathbb{R}^{n \times n} \ni \b{\Delta} := \b{\Omega}_S^{(1/2)} \b{\Psi}_S^\top$.
The $\b{\Delta}\b{U}$ can be interpreted as manipulated (or rotated) projection directions.
We want the manipulated projection matrix to be orthogonal; thus:
\begin{align}\label{equation_orthogonal_manipulated_projection}
(\b{\Delta}\b{U})^\top (\b{\Delta}\b{U}) = \b{U}^\top \b{S}_2\, \b{U} \overset{\text{set}}{=} \b{I}.
\end{align}
In this case, the optimization is expressed as:
\begin{equation}\label{equation_optimization_generalForm_generalized_eigenvalue}
\begin{aligned}
& \underset{\b{U}}{\text{maximize}}
& & \textbf{tr}(\b{U}^\top \b{S}_1\, \b{U}), \\
& \text{subject to}
& & \b{U}^\top \b{S}_2\, \b{U} = \b{I},
\end{aligned}
\end{equation}
and the constraint makes the problem well-defined.
The Lagrangian \cite{boyd2004convex} of the problem is:
\begin{align*}
\mathcal{L} = \textbf{tr}(\b{U}^\top \b{S}_1\, \b{U}) - \textbf{tr}\big(\b{\Lambda}^\top (\b{U}^\top \b{S}_2\, \b{U} - \b{I})\big),
\end{align*}
where $\b{\Lambda}$ is a diagonal matrix which includes the Lagrange multipliers. 
Setting the derivative of Lagrangian to zero gives:
\begin{align}
& \frac{\partial \mathcal{L}}{\partial \b{U}} = 2\b{S}_1 \b{U} - 2\b{S}_2\b{U} \b{\Lambda} \overset{\text{set}}{=} \b{0} \implies \b{S}_1\, \b{U} = \b{S}_2\, \b{U} \b{\Lambda}, \label{equation_scatter_generalized_eigendecomposition}
\end{align}
which is the generalized eigenvalue problem $(\b{S}_1, \b{S}_2)$ where the columns of $\b{U}$ and the diagonal of $\b{\Lambda}$ are the eigenvectors and eigenvalues, respectively \cite{ghojogh2019eigenvalue}. The eigenvectors and eigenvalues are again sorted from the leading to the trailing because of maximization.

\section{PCA, FDA, and SPCA}\label{section_PCA_FDA_SPCA}

The optimization problem in PCA \cite{pearson1901liii,jolliffe2011principal,ghojogh2019unsupervised} is expressed as:
\begin{equation}\label{equation_optimization_PCA}
\begin{aligned}
& \underset{\b{U}}{\text{maximize}}
& & \textbf{tr}(\b{U}^\top \b{S}_T\, \b{U}), \\
& \text{subject to}
& & \b{U}^\top \b{U} = \b{I},
\end{aligned}
\end{equation}
where $\b{S}_T \in \mathbb{R}^{n \times n}$ is the total scatter defined in Eq. (\ref{equation_total_scatter}).
The solution to Eq. (\ref{equation_optimization_PCA}) is the eigenvalue problem for $\b{S}_T$ according to Eq. (\ref{equation_scatter_eigendecomposition}).
Thus, the PCA directions are the eigenvectors of the total scatter. 

The FDA \cite{fisher1936use,friedman2001elements} maximizes the Fisher criterion \cite{xu2006analysis,fukunaga2013introduction}:
\begin{align}\label{equation_optimization_FDA_criterion}
&\underset{\b{U}}{\text{maximize}} ~~~ f_F(\b{U}) := \frac{d_B(\b{U})}{d_W(\b{U})} :=  \frac{\textbf{tr}(\b{U}^\top \b{S}_B\, \b{U})}{\textbf{tr}(\b{U}^\top \b{S}_W\, \b{U})}. 
\end{align}
The Fisher criterion $f_F(\b{U})$ is a generalized Rayleigh-Ritz Quotient \cite{parlett1998symmetric}.
Hence, the optimization in Eq. (\ref{equation_optimization_FDA_criterion}) is equivalent to \cite{ghojogh2019fisher}:
\begin{equation}\label{equation_optimization_FDA}
\begin{aligned}
& \underset{\b{U}}{\text{maximize}}
& & \textbf{tr}(\b{U}^\top \b{S}_B\, \b{U}), \\
& \text{subject to}
& & \b{U}^\top \b{S}_W\, \b{U} = \b{I},
\end{aligned}
\end{equation}
where the $\b{S}_B$ and $\b{S}_W$ are the between and within scatters, respectively, defined as:
\begin{align}
&\mathbb{R}^{d \times d} \ni \b{S}_B := \sum_{j=1}^c n_j (\b{\mu}_j - \b{\mu}) (\b{\mu}_j - \b{\mu})^\top, \label{equation_between_scatter} \\
&\mathbb{R}^{d \times d} \ni \b{S}_W := \sum_{j=1}^c \sum_{i=1}^{n_j} (\b{x}_i^{(j)} - \b{\mu}_j) (\b{x}_i^{(j)} - \b{\mu}_j)^\top, \label{equation_within_scatter}
\end{align}
where the mean of the $j$-th class is:
\begin{align}\label{equation_mean_of_class}
\mathbb{R}^{t} \ni \b{\mu}_j := \frac{1}{n_j} \sum_{i=1}^{n_j} \b{x}_i^{(j)}.
\end{align}

The total scatter can be considered as the summation of the between and within scatters \cite{ye2007least,welling2005fisher}:
\begin{align}\label{equation_S_T_as_sum_of_scatters}
\b{S}_T = \b{S}_B + \b{S}_W \implies \b{S}_B = \b{S}_T - \b{S}_W.
\end{align}
Therefore, the Fisher criterion can be written as \cite{welling2005fisher}:
\begin{align}\label{equation_optimization_FDA_criterion_with_S_T}
&f_F(\b{U}) = \frac{d_T(\b{U})}{d_W(\b{U})} - 1 := \frac{\textbf{tr}(\b{U}^\top \b{S}_T\, \b{U})}{\textbf{tr}(\b{U}^\top \b{S}_W\, \b{U})} - 1. 
\end{align}
The $-1$ is a constant and can be dropped in the optimization problem because the variable $\b{U}$ and not the objective is the goal; therefore, the optimization in FDA can now be expressed as:
\begin{equation}\label{equation_optimization_FDA_with_S_T}
\begin{aligned}
& \underset{\b{U}}{\text{maximize}}
& & \textbf{tr}(\b{U}^\top \b{S}_T\, \b{U}), \\
& \text{subject to}
& & \b{U}^\top \b{S}_W\, \b{U} = \b{I}.
\end{aligned}
\end{equation}
Hence, the FDA directions can be obtained by the generalized eigenvalue problem $(\b{S}_T, \b{S}_W)$ \cite{welling2005fisher}.
Note that some articles, such as \cite{ye2007least,zhang2010regularized,diaz2017deep}, solve the generalized eigenvalue problem $(\b{S}_B, \b{S}_T)$ by considering another version of the Fisher criterion which is $\textbf{tr}(\b{U}^\top \b{S}_B\, \b{U}) / \textbf{tr}(\b{U}^\top \b{S}_T\, \b{U})$. This criterion is obtained if we consider minimization of the inverse of Eq. (\ref{equation_optimization_FDA_criterion}), use Eq. (\ref{equation_S_T_as_sum_of_scatters}) for $\b{S}_W$, drop the constant $-1$, and convert minimization to maximization by inverting the criterion again.

Comparing the Eqs. (\ref{equation_optimization_PCA}) and (\ref{equation_optimization_FDA_with_S_T}) shows that PCA captures the orthonormal directions with the maximum variance of data; however, FDA has the same goal but also it requires the manipulated directions to be orthonormal. This manipulation is done by the within scatter which makes sense because the within scatters make use of the class labels. 
This comparison gives a hint for the connection between PCA and FDA.


The SPCA \cite{barshan2011supervised} makes use of the empirical estimation of the Hilbert-Schmidt Independence Criterion (HSIC) \cite{gretton2005measuring}:
\begin{align}\label{equation_HSIC}
\text{HSIC} := \frac{1}{(n-1)^2}\, \textbf{tr}(\b{K}_1 \b{H}\b{K}_2 \b{H}),
\end{align}
where $\b{K}_1$ and $\b{K}_2$ are the kernels over the first and second random variable. The idea of HSIC is to measure the dependence of two random variables by calculating the correlation of their pulled values to the Hilbert space. 
The SPCA uses HSIC for the projected data $\b{U}^\top \b{X}$ and the labels $\b{Y}$ and maximizes the dependence of them in order to make use of the labels. It uses the linear kernel for the projected data $\b{K}_1 = (\b{U}^\top \b{X})^\top (\b{U}^\top \b{X}) = \b{X}^\top \b{U} \b{U}^\top \b{X}$ and an arbitrary valid kernel for the labels $\b{K}_2=\b{K}_y$. Therefore, the scaled Eq. (\ref{equation_HSIC}) in SPCA is:
\begin{align}\label{equation_HSIC_for_SPCA}
\textbf{tr}(\b{X}^\top \b{U} \b{U}^\top \b{X} \b{H}\b{K}_y \b{H}) \overset{(a)}{=} \textbf{tr}(\b{U}^\top \b{X} \b{H} \b{K}_y \b{H} \b{X}^\top \b{U}),
\end{align}
where $(a)$ is because of the cyclic property of the trace. 
The optimization problem in SPCA is expressed as:
\begin{equation}\label{equation_optimization_SPCA}
\begin{aligned}
& \underset{\b{U}}{\text{maximize}}
& & \textbf{tr}(\b{U}^\top \b{X} \b{H} \b{K}_y \b{H} \b{X}^\top \b{U}), \\
& \text{subject to}
& & \b{U}^\top \b{U} = \b{I},
\end{aligned}
\end{equation}
where $\b{K}_y$ is the kernel matrix over the labels of data, either for classification or regression. 
The solution to Eq. (\ref{equation_optimization_SPCA}) is the eigenvalue problem for $\b{X}\b{H}\b{K}_y\b{H}\b{X}^\top$ according to Eq. (\ref{equation_scatter_eigendecomposition}).
Hence, the SPCA directions are the eigenvectors of $\b{X}\b{H}\b{K}_y\b{H}\b{X}^\top$. 
Note that this term, restricted to a linear kernel for $\b{K}_y$, is used as the between scatter in \cite{zhang2010regularized,diaz2017deep,diaz2019deep} hinting for a connection between SPCA and FDA if we compare the objectives in Eqs. (\ref{equation_optimization_FDA}) and (\ref{equation_optimization_SPCA}).

PCA, FDA, and SPCA, which were explained, are several fundamental subspace learning methods which are different in terms of whether/how to use the labels. Comparing Eqs. (\ref{equation_optimization_PCA}), (\ref{equation_optimization_FDA_with_S_T}), and (\ref{equation_optimization_SPCA}) and considering the general forms in Eqs. (\ref{equation_optimization_generalForm_eigenvalue}) and (\ref{equation_optimization_generalForm_generalized_eigenvalue}) show that these methods belong to a family of methods based on eigenvalue and generalized eigenvalue problems. This gave us a motivation to propose a generalized subspace learning method, named RDA, as a family of methods including PCA, FDA, and SPCA. 

\section{Roweis Discriminant Analysis (RDA)}\label{section_RDA}

\subsection{Methodology}

RDA aims at maximizing the $\textbf{tr}(\b{U}^\top \b{R}_1\, \b{U})$ interpreted as the squared length of the reconstruction or the scatter of projection (see Eq. (\ref{equation_squared_length_reconstruction})), while requiring the manipulated projection directions to be orthonormal (see Eq. (\ref{equation_orthogonal_manipulated_projection})). Therefore, the optimization of RDA is formalized as:
\begin{equation}\label{equation_optimization_RDA}
\begin{aligned}
& \underset{\b{U}}{\text{maximize}}
& & \textbf{tr}(\b{U}^\top \b{R}_1\, \b{U}), \\
& \text{subject to}
& & \b{U}^\top \b{R}_2\, \b{U} = \b{I},
\end{aligned}
\end{equation}
where $\b{R}_1$ and $\b{R}_2$ are the first and second \textit{Roweis matrices} defined as:
\begin{align}
&\mathbb{R}^{d \times d} \ni \b{R}_1 := \b{X} \b{H} \b{P} \b{H} \b{X}^\top, \label{equation_R1} \\
&\mathbb{R}^{d \times d} \ni \b{R}_2 := r_2\, \b{S}_W + (1 - r_2)\, \b{I}, \label{equation_R2}
\end{align}
respectively, where:
\begin{align}\label{equation_P}
\mathbb{R}^{n \times n} \ni \b{P} := r_1\, \b{K}_y + (1 - r_1)\, \b{I}.
\end{align}
The $r_1 \in [0,1]$ and $r_2 \in [0,1]$ are the first and second \textit{Roweis factors}. 
Note that $\b{H}\b{P}\b{H}$ in Eq. (\ref{equation_R1}) is double-centering the matrix $\b{P}$.

The solution to Eq. (\ref{equation_optimization_RDA}) is the generalized eigenvalue problem $(\b{R}_1, \b{R}_2)$ according to Eq. (\ref{equation_scatter_generalized_eigendecomposition}). Therefore, the RDA directions are the eigenvectors of this generalized eigenvalue problem. The RDA directions are sorted from the leading to trailing eigenvalues because of the maximization in Eq. (\ref{equation_optimization_RDA}). 

The optimization of the RDA can be interpreted in another way using what we define as the \textit{Roweis criterion}. We want to maximize this criterion:
\begin{align}\label{equation_optimization_RDA_criterion}
&\underset{\b{U}}{\text{maximize}} ~~~ f_R(\b{U}) := \frac{d_{R_1}(\b{U})}{d_{R_2}(\b{U})} := \frac{\textbf{tr}(\b{U}^\top \b{R}_1\, \b{U})}{\textbf{tr}(\b{U}^\top \b{R}_2\, \b{U})}. 
\end{align}
As in FDA, the Roweis criterion is a generalized Rayleigh-Ritz Quotient \cite{parlett1998symmetric}; thus, the optimization in Eq. (\ref{equation_optimization_RDA_criterion}) is equivalent to Eq. (\ref{equation_optimization_RDA}).
It is noteworthy that if we consider only a one-dimensional RDA subspace, the Roweis criterion is:
\begin{align}\label{equation_RDA_criterion_oneDimensional}
f_R(\b{u}) := \frac{d_{R_1}(\b{u})}{d_{R_2}(\b{u})} := \frac{\b{u}^\top \b{R}_1\, \b{u}}{\b{u}^\top \b{R}_2\, \b{u}},
\end{align}
where $\b{u} \in \mathbb{R}^d$ is the only RDA projection direction.
In this case, the Eq. (\ref{equation_optimization_RDA}) becomes:
\begin{equation}\label{equation_optimization_RDA_ondimensional}
\begin{aligned}
& \underset{\b{u}}{\text{maximize}}
& & \b{u}^\top \b{R}_1\, \b{u}, \\
& \text{subject to}
& & \b{u}^\top \b{R}_2\, \b{u} = 1.
\end{aligned}
\end{equation}

\subsection{The Special Cases of RDA \& the Roweis Map}

Consider the Eqs. (\ref{equation_R1}), (\ref{equation_R2}), and (\ref{equation_P}). 
We know that the range of both $r_1$ and $r_2$ is $[0,1]$. If we consider the extreme cases of $r_1$ and $r_2$, we find an interesting relationship:
\begin{align}
&r_1=0,\, r_2=0 \implies \text{RDA} \equiv \text{PCA}, \\
&r_1=0,\, r_2=1 \implies \text{RDA} \equiv \text{FDA}, \\
&r_1=1,\, r_2=0 \implies \text{RDA} \equiv \text{SPCA},
\end{align}
by comparing Eq. (\ref{equation_optimization_RDA}) with Eqs. (\ref{equation_optimization_PCA}), (\ref{equation_optimization_FDA_with_S_T}), and (\ref{equation_optimization_SPCA}) and noticing the Eq. (\ref{equation_total_scatter}).
We see that PCA, FDA, and SPCA are all special cases of RDA.

\begin{figure}[!t]
\centering
\includegraphics[width=2.5in]{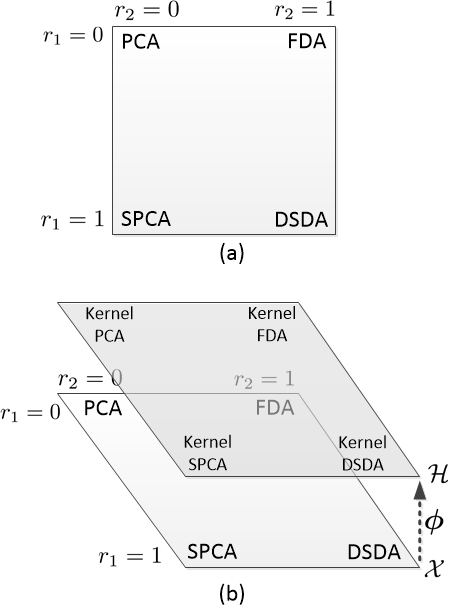}
\caption{The Roweis map: (a) the map including infinite number of subspace learning methods and its special cases, (b) the map in its input and feature spaces where the map in the feature space is pulled from the map in the input space.}
\label{figure_Roweis_map}
\end{figure}

In fact, RDA is a family of infinite number of algorithms for subspace learning. By choosing any number for the $r_1$ and $r_2$ in the range $[0,1]$, RDA gives us a new algorithm for learning the subspace of data. 
We define a map, named \textit{Roweis map}, which includes the infinite number of special cases of RDA where three of its corners are PCA, FDA, and SPCA. The rows and columns of the Roweis map are the values of $r_1$ and $r_2$, respectively. 
Figure \ref{figure_Roweis_map}-a shows this map.

An interesting thing about RDA is that one of the four extreme values for $r_1$ and $r_2$ is not yet proposed in the literature to the best of our knowledge. If $r_1=1$ and $r_2=1$, the Eq. (\ref{equation_optimization_RDA}) becomes:
\begin{equation}\label{equation_optimization_DSDA}
\begin{aligned}
& \underset{\b{U}}{\text{maximize}}
& & \textbf{tr}(\b{U}^\top \b{X}\b{H}\b{K}_y\b{H}\b{X}^\top\, \b{U}), \\
& \text{subject to}
& & \b{U}^\top \b{S}_W\, \b{U} = \b{I},
\end{aligned}
\end{equation}
whose solution is the generalized eigenvalue problem $(\b{X}\b{H}\b{K}_y\b{H}\b{X}^\top, \b{S}_W)$ according to Eq. (\ref{equation_scatter_generalized_eigendecomposition}).
As can be seen, this optimization uses the labels twice, once in the kernel over the labels and once in the within scatter. Therefore, we name this special subspace learning method the \textit{Double Supervised Discriminant Analysis (DSDA)}.

\subsection{Properties of the Roweis Matrices and Factors}

\subsubsection{Properties of the Roweis Factors}

\begin{figure}[!t]
\centering
\includegraphics[width=2.3in]{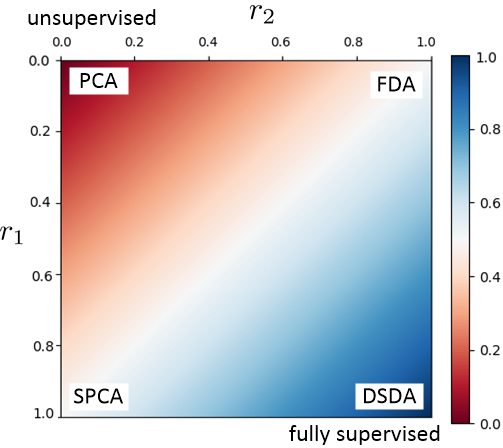}
\caption{The supervision level shown on the Roweis map (best viewed in color).}
\label{figure_s_plot}
\end{figure}

In Eq. (\ref{equation_P}), if $r_1=1$, we are fully using the kernel over labels and if $r_1=0$, it reduces to the identity matrix. So, we have:
\begin{align}\label{equation_P_bracket}
\b{P} = 
\left\{
    \begin{array}{ll}
        \b{K}_y & \text{if } r_1=1, \\
        \b{I} & \text{if } r_1=0.
    \end{array}
\right.
\end{align}
On the other hand, in Eq. (\ref{equation_R2}), if $r_2=1$, we are fully using the within scatter using the labels and if $r_1=0$, it reduces to the identity matrix. Hence:
\begin{align}\label{equation_R2_bracket}
\b{R}_2 = 
\left\{
    \begin{array}{ll}
        \b{S}_W & \text{if } r_2=1, \\
        \b{I} & \text{if } r_2=0.
    \end{array}
\right.
\end{align}
Therefore, we can conclude that if the Roweis factor is one, we fully use the labels as a supervised method and if the Roweis factor is zero, we are not using the labels at all. Therefore, the Roweis factor is a measure of using labels or being supervised. As we have two Roweis factors, we define:
\begin{align}\label{equation_s}
[0,1] \ni s := (r_1 + r_2) / 2,
\end{align}
as the \textit{supervision level} which is a planar function depicted in Fig. \ref{figure_s_plot}. Note that the extremes $s=0$ and $s=1$ refer to the unsupervised and fully (double) supervised subspace learning, respectively. Recall that the Roweis map includes the different subspace learning methods from unsupervised PCA (top-left corner of map) to the DSDA (bottom-right corner of map). The value $s=0.5$ can be interpreted as using the labels once (as in both FDA and SPCA). 

It is noteworthy that if $r_2=0$, we have $\b{R}_2=\b{I}$ according to Eq. (\ref{equation_R2_bracket}). Therefore, according to Eqs. (\ref{equation_optimization_RDA}) and (\ref{equation_P_bracket}), the optimization of RDA merely includes $\b{K}_y$ and not $\b{S}_W$ regarding the labels. On the other hand, the $\b{K}_y$ is a soft measure of similarity between the labels while the $\b{S}_W$ uses the labels strictly for knowing which data instance belongs to which class. It shows that if we use $\b{S}_W$ ($r_2 \neq 0$), the labels must be for classes in classification; however, if using solely $\b{K}_y$ ($r_2 = 0$), the labels can be either for classification or regression. This sheds light to why SPCA (with $r_2=0$) can be used for both classification and regression tasks \cite{barshan2011supervised} but FDA (with $r_2=1$) is only for classification.

\subsubsection{Properties of the Roweis Matrices}

The Roweis matrices, $\b{R}_1$ and $\b{R}_2$, are positive semi-definite. The reason is that the within scatter, the kernel matrix, and the identity matrix are positive semi-definite and $r_1, r_2 \in [0,1]$, so:
\begin{align}
&r_1\, \b{K}_y \succeq 0,\, (1 - r_1)\b{I} \succeq 0 \implies \b{P} \succeq 0,\, \b{R}_1 \succeq 0, \label{equation_R1_positive_semidefinite} \\
&r_2\, \b{S}_W \succeq 0,\, (1 - r_2)\b{I} \succeq 0 \implies \b{R}_2 \succeq 0. \label{equation_R2_positive_semidefinite}
\end{align}
The Roweis matrices are also symmetric because the within scatter, the kernel matrix, and the identity matrix are symmetric; hence:
\begin{align}\label{equation_Roweis_matrices_symmetric}
\b{R}_1 = \b{R}_1^\top, ~~~ \b{R}_2 = \b{R}_2^\top, ~~~ \b{P} = \b{P}^\top.
\end{align}

It is also worth mentioning that according to the definitions of $\b{R}_1$, $\b{R}_2$, and $\b{P}$, the $\b{P}$ and the second Roweis matrices have the essence of kernel and scatter matrices, respectively, while the first Roweis matrix has the mixture essence of scatter and kernel matrices.
According to Eq. (\ref{equation_HSIC_for_SPCA}), the first Roweis matrix is the term used in the HSIC over the projected data and the labels. 

Now, we analyze the rank of the Roweis matrices.
We have $\mathbb{R}^{n \times n} \ni \b{K}_y := \b{\Phi}(\b{Y})^\top \b{\Phi}(\b{Y})$ where $\b{\Phi}(\b{Y}) \in \mathbb{R}^t$ is the pulled $\b{Y}$ to the feature space \cite{hofmann2008kernel}. We usually have $t \gg \ell$. Hence, $\textbf{rank}(\b{K}_y) \leq \min(n,t)$. According to Eq. (\ref{equation_P}) and the subadditivity property of the rank:
\begin{align}\label{equation_rank_P}
\textbf{rank}(\b{P}) \leq \textbf{rank}(\b{K}_y) + \textbf{rank}(\b{I}_n) \leq \min(n,t) + n.
\end{align}
Note that in the extreme cases $r_1=0$ and $r_1=1$, the rank of $\b{P}$ is $n$ and $\leq \min(n,t)$, respectively. 
According to Eq. (\ref{equation_R1}):
\begin{align}
\textbf{rank}(\b{R}_1) &\leq \min\big( \textbf{rank}(\b{X}) + \textbf{rank}(\b{H}) + \textbf{rank}(\b{P}) \big) \nonumber \\
&\leq \min(d,n-1,t) = \min(d,n-1), \label{equation_rank_R1}
\end{align}
where the $-1$ is because of subtracting the mean in centering the matrix (e.g., consider the case with only one data instance). 
In general, the rank of a covariance (scatter) matrix over the $d$-dimensional data with sample size $n$ is at most $\min(d, n-1)$. The $d$ is because the covariance matrix is a $d \times d$ matrix, the $n$ is because we iterate over $n$ data instances for calculating the covariance matrix, and the $-1$ is again because of subtracting the mean.
Hence, according to Eq. (\ref{equation_within_scatter}), we have $\textbf{rank}(\b{S}_W) \leq \min(d, n-1)$.
According to Eq. (\ref{equation_R2}) and the subadditivity property of the rank:
\begin{align}\label{equation_rank_R2}
\textbf{rank}(\b{R}_2) \leq \textbf{rank}(\b{S}_W) + \textbf{rank}(\b{I}_d) \leq \min(d,n-1) + d.
\end{align}
In the extreme cases $r_2=0$ and $r_2=1$, the rank of $\b{R}_2$ is $d$ and $\leq \min(d, n-1)$, respectively. 

\subsection{Dimensionality of the RDA Subspace}

There exist a rigorous solution for the generalized eigenvalue problem $(\b{R}_1, \b{R}_2)$ \cite{ghojogh2019eigenvalue}. However, we can solve this problem as:
\begin{align}
\b{R}_1\, \b{U} = \b{R}_2\, \b{U} \b{\Lambda} &\implies \b{R}_2^{-1} \b{R}_1\, \b{U} = \b{U} \b{\Lambda} \nonumber \\
&\implies \b{U} = \textbf{eig}(\b{R}_2^{-1} \b{R}_1), \label{equation_RDA_dirty_solution}
\end{align}
where $\textbf{eig}(.)$ stacks the eigenvectors column-wise. 
If $\b{R}_2$ is singular, we strengthen its diagonal:
\begin{align}
\b{U} = \textbf{eig}\big((\b{R}_2 + \varepsilon \b{I})^{-1} \b{R}_1\big), \label{equation_RDA_dirty_solution_singular}
\end{align}
where $\varepsilon$ is a very small positive number, large enough to make $\b{R}_2$ full rank. In the literature, this approach is known as regularized discriminant analysis \cite{friedman1989regularized}.

According to Eqs. (\ref{equation_rank_R1}) and (\ref{equation_rank_R2}), the rank of $\b{R}_2^{-1} \b{R}_1$ in Eq. (\ref{equation_RDA_dirty_solution}) is:
\begin{align}
\textbf{rank}(\b{R}_2^{-1} \b{R}_1) &\leq  \min\big(\textbf{rank}(\b{R}_2^{-1}), \textbf{rank}(\b{R}_1)\big) \nonumber \\
&\leq \min(d,n-1).
\end{align}
Therefore, the $p$, which is the dimensionality of the RDA subspace, is at most $\min(d,n-1)$. The $\min(d,n-1)$ leading eigenvectors are considered as the RDA directions and the rest of eigenvectors are invalid and ignored for having zero or very small eigenvalues. 

It is noteworthy that according to Eq. (\ref{equation_between_scatter}), we have $\textbf{rank}(\b{S}_B) \leq \min(d,c-1)=c-1$. If Eq. (\ref{equation_optimization_FDA}) is used for FDA, we have $p \leq c-1$ in FDA. 
In RDA, for the values $r_1=0, r_2=1$, RDA is reduced to FDA when the Eq. (\ref{equation_optimization_FDA_with_S_T}) is used; hence, the dimensionality of FDA as a special case of RDA is at most $\min(d,n-1)$ and not $c-1$.

\subsection{Robust RDA}


According to Eq. (\ref{equation_rank_R2}), the rank of $\b{R}_2 \in \mathbb{R}^{d \times d}$ is at most $\min(d,n-1) + d$.
In the extreme case $r_2=1$, its rank is at most $\min(d,n-1)$.
Thus, for the case $r_2$ is very close to one, i.e., $r_2 \approx 1$, if $n-1<d$, the $\b{R}_2$ is singular. This happens when the dimensionality of data is huge but the sample size is small. 
In this case, we face a problem if we use the Eq. (\ref{equation_RDA_dirty_solution}), for solving the generalized eigenvalue problem $(\b{R}_1, \b{R}_2)$.
Two possible approaches to tackle this problem are the pseudo-inverse of $\b{R}_2$ \cite{webb2003statistical} and strengthening the diagonal of $\b{R}_2$ \cite{friedman1989regularized,diaz2017deep}.
We can also propose the Robust RDA (RRDA) inspired by robust FDA \cite{deng2007robust,guo2015feature} to be robust against singularity. 
We emphasis that RRDA is useful if $r_2 \approx 1$, $n-1<d$, where we want to use Eq. (\ref{equation_RDA_dirty_solution}) for solving the generalized eigenvalue problem. 

In RRDA, the $\b{R}_2$ is decomposed using eigenvalue decomposition: $\b{R}_1 = \b{\Phi}^\top \b{\Lambda} \b{\Phi}$ where $\b{\Phi}$ and $\b{\Lambda} = \textbf{diag}([\lambda_1, \dots, \lambda_d]^\top)$ include the eigenvectors and eigenvalues of $\b{R}_1$, respectively. 
The eigenvalues are sorted as $\lambda_1 \geq \dots \geq \lambda_d$ and the eigenvectors are sorted accordingly. 
If $\b{R}_1$ is close to singularity, the first $d'$ eigenvalues are valid and the rest $(d-d')$ eigenvalues are either very small or zero. 
The appropriate $d'$ is obtained as:
\begin{align}
d' := \arg\min_{m} \bigg(\frac{\sum_{j=1}^m \lambda_j}{\sum_{k=1}^d \lambda_k} \geq 0.98 \bigg),
\end{align}
where $0.98$ is a selected number close to $1$.
The $(d - d')$ invalid eigenvalues are replaced with $\lambda_*$:
\begin{align}
\mathbb{R}^{d \times d} \ni \b{\Lambda}' := \textbf{diag}([\lambda_1, \dots, \lambda_{d'}, \lambda_*, \dots, \lambda_*]^\top),
\end{align}
where \cite{deng2007robust}:
\begin{align}
\lambda_* := \frac{1}{d - d'} \sum_{j=d'+1}^d \lambda_j.
\end{align}
Hence, the $\b{R}_2$ is replaced with $\b{R}'_2$:
\begin{align}
\mathbb{R}^{d \times d} \ni \b{R}'_2 := \b{\Phi}^\top \b{\Lambda}' \b{\Phi},
\end{align}
and the robust RDA directions are the eigenvectors of the generalized eigenvalue problem $(\b{R}_1, \b{R}'_2)$.

\section{Dual RDA for $r_2 = 0$}\label{section_dual_RDA}

The dual RDA exists for $r_2=0$ for the reason explained later in this section. This statement clarifies why, in the literature, PCA and SPCA (with $r_2=0$) have their dual methods but FDA (with $r_2=1$) does not have a dual.

As the matrix $\b{K}_y$ is symmetric and positive semi-definite, we can decompose it as:
\begin{align}
\b{K}_y \overset{(a)}{=} \b{\Psi}_K\, \b{\Omega}_K \b{\Psi}_K^\top = \b{\Psi}_K\, \b{\Omega}_K^{(1/2)} \b{\Omega}_K^{(1/2)} \b{\Psi}_K^\top \overset{(b)}{=} \b{\Upsilon} \b{\Upsilon}^\top, \label{equation_K_y_decompose}
\end{align}
where $(a)$ is because of SVD and $(b)$ is for:
\begin{align}
\mathbb{R}^{n \times n} \ni \b{\Upsilon} := \b{\Psi}_K\, \b{\Omega}_K^{(1/2)}.
\end{align}
Thus, the $\b{R}_1$ can be decomposed as:
\begin{align}
\b{R}_1 &= \b{X} \b{H} \big(r_1\, \b{K}_y + (1 - r_1)\, \b{I}\big) \b{H} \b{X}^\top \nonumber \\
&\overset{(a)}{=} r_1\, \b{X} \b{H} \b{\Upsilon} \b{\Upsilon}^\top \b{H}^\top \b{X}^\top + (1 - r_1)\, \b{X} \b{H} \b{H}^\top \b{X}^\top \nonumber \\
&\overset{(b)}{=} r_1\, \b{Q} \b{Q}^\top + (1 - r_1)\, \breve{\b{X}} \breve{\b{X}}^\top \nonumber \\
&\overset{(\ref{equation_total_scatter})}{=} r_1\, \b{Q} \b{Q}^\top + (1 - r_1)\, \b{S}_T \overset{(c)}{=} \b{W} \b{W}^\top, \label{equation_R1_decompose_1}
\end{align}
where $(a)$ is because of Eq. (\ref{equation_K_y_decompose}) and $\b{H}$ is symmetric, $(b)$ is because of Eq. (\ref{equation_centered_training_data}) and:
\begin{align}
\mathbb{R}^{d \times n} \ni \b{Q} := \b{X} \b{H} \b{\Upsilon},
\end{align}
and $(c)$ is because the $\b{Q}\b{Q}^\top$ and $\breve{\b{X}} \breve{\b{X}}^\top$ are both symmetric and in the the same form so their summation is also symmetric (recall Eq. (\ref{equation_Roweis_matrices_symmetric})) and in the same form.
We can obtain Eq. (\ref{equation_R1_decompose_1}) in another way, too:
as $\b{R}_1$ is symmetric and positive semi-definite (see Eq. (\ref{equation_R1_positive_semidefinite})), we can decompose it:
\begin{align}
\b{R}_1 \overset{(a)}{=} \b{\Psi}_R\, \b{\Omega}_R\, \b{\Psi}_R^\top = \b{\Psi}_R\, \b{\Omega}_R^{(1/2)} \b{\Omega}_R^{(1/2)} \b{\Psi}_R^\top \overset{(b)}{=} \b{W} \b{W}^\top, \label{equation_R1_decompose_2}
\end{align}
where $(a)$ is because of SVD and $(b)$ is for:
\begin{align}
\b{W} := \b{\Psi}_R\, \b{\Omega}_R^{(1/2)}. \label{equation_W}
\end{align}
For the reason that will be explained at the end of this section, we use the incomplete SVD in $(a)$ for $\b{R}_1$ where $\b{\Psi}_R \in \mathbb{R}^{d \times k}$, $\b{\Omega}_R \in \mathbb{R}^{k \times k}$, $k := \min(d,n)$ \cite{golub1971singular}. Therefore, we have $\b{W} \in \mathbb{R}^{d \times k}$. If we had used the complete SVD, we would have $\b{W} \in \mathbb{R}^{d \times d}$.

If $r_2 = 0$, we have $\b{R}_2 = \b{I}$ according to Eq. (\ref{equation_R2}). In this case, the Eq. (\ref{equation_optimization_RDA}) becomes similar to Eq. (\ref{equation_optimization_generalForm_eigenvalue}) and thus the solution which is the generalized eigenvalue problem $(\b{R}_1, \b{R}_2)$ gets reduced to the eigenvalue problem for $\b{R}_1$. 
According to the properties of SVD, the matrix of left singular vectors of $\b{W}$ is equivalent to the matrix of eigenvectors of $\b{W}\b{W}^\top = \b{R}_1$. Therefore, we can use incomplete SVD for $\b{W}$:
\begin{align}\label{equation_W_SVD}
\mathbb{R}^{d \times k} \ni \b{W} = \b{U} \b{\Sigma} \b{V}^\top,
\end{align}
where the columns of $\b{U} \in \mathbb{R}^{d \times k}$ and $\b{V} \in \mathbb{R}^{d \times k}$ are the left and right singular vectors (i.e., the eigenvectors of $\b{W}\b{W}^\top$ and $\b{W}^\top\b{W}$), respectively. The diagonal entries of $\b{\Sigma} \in \mathbb{R}^{k \times k}$ are the singular value of $\b{W}$, i.e., the square root of eigenvalues of $\b{W}\b{W}^\top$ or $\b{W}^\top\b{W}$.

The dual RDA exists only for $r_2=0$ because in Eq. (\ref{equation_W_SVD}), the $\b{U}$ is an orthogonal matrix so $\b{U}^\top \b{U} = \b{I}$. This implies that the constraint in Eq. (\ref{equation_optimization_RDA}) should be $\b{U}^\top \b{U} = \b{I}$ which means $\b{R}_2 = \b{I}$. Recall that this is the reason for FDA (with $r_2 \neq 0$) not having a dual.

We obtain $\b{U}$ from Eq. (\ref{equation_W_SVD}):
\begin{align}
\b{W}\b{V} \overset{(\ref{equation_W_SVD})}{=} \b{U}\b{\Sigma} \implies \b{U} = \b{W}\b{V}\b{\Sigma}^{-1}, \label{equation_U_dual_RDA}
\end{align}
where the orthogonality of $\b{V}$ is noticed. 
The projection of data $\b{X}$ in dual RDA is:
\begin{align}
\widetilde{\b{X}} &\overset{(\ref{equation_projection_severalPoint})}{=} \b{U}^\top \b{X} \overset{(\ref{equation_U_dual_RDA})}{=} (\b{W}\b{V} \b{\Sigma}^{-1})^\top \b{X} = \b{\Sigma}^{-\top}\b{V}^\top\b{W}^\top\b{X} \nonumber \\
& \overset{(\ref{equation_W})}{=} \b{\Sigma}^{-1}\b{V}^\top \b{\Omega}_R^{(1/2)\top} \b{\Psi}_R^\top\, \b{X}. \label{equation_projected_dual_RDA}
\end{align}
Note that $\b{\Sigma}$ is symmetric. 
Similarly, out-of-sample projection in dual RDA is:
\begin{align}\label{equation_outOfSample_projected_dual_RDA}
\widetilde{\b{X}}_t = \b{\Sigma}^{-1}\b{V}^\top \b{\Omega}_R^{(1/2)\top} \b{\Psi}_R^\top\, \b{X}_t.
\end{align}

The reconstruction of training data in dual RDA is:
\begin{align}
\widehat{\b{X}} &\overset{(\ref{equation_reconstruction_severalPoint})}{=} \b{U}\b{U}^\top \b{X} = \b{U}\widetilde{\b{X}} \nonumber \\
&\overset{(a)}{=} \b{W}\b{V}\b{\Sigma}^{-1} \b{\Sigma}^{-1}\b{V}^\top \b{\Omega}_R^{(1/2)\top} \b{\Psi}_R^\top\, \b{X} \nonumber \\
&\overset{(\ref{equation_W})}{=} \b{\Psi}_R\, \b{\Omega}_R^{(1/2)}\b{V}\b{\Sigma}^{-2} \b{V}^\top \b{\Omega}_R^{(1/2)\top} \b{\Psi}_R^\top\, \b{X}, \label{equation_reconstruction_dual_SPCA}
\end{align}
where $(a)$ is because of Eqs. (\ref{equation_U_dual_RDA}) and (\ref{equation_projected_dual_RDA}).
Similarly, reconstruction of out-of-sample data in dual RDA is:
\begin{align}
\widehat{\b{X}}_t = \b{\Psi}_R\, \b{\Omega}_R^{(1/2)}\b{V}\b{\Sigma}^{-2} \b{V}^\top \b{\Omega}_R^{(1/2)\top} \b{\Psi}_R^\top\, \b{X}_t. \label{equation_outOfSample_reconstruction_dual_SPCA}
\end{align}

To summarize, in RDA, the projection of training and out-of-sample data is Eq. (\ref{equation_projection_severalPoint}) and (\ref{equation_outOfSample_projection_severalPoint}), respectively. 
In RDA, the reconstruction of training and out-of-sample data is Eq. (\ref{equation_reconstruction_severalPoint}) and (\ref{equation_outOfSample_reconstruction_severalPoint}), respectively. 
However, in dual RDA, the projection of training and out-of-sample data is Eq. (\ref{equation_projected_dual_RDA}) and (\ref{equation_outOfSample_projected_dual_RDA}), respectively. 
In dual RDA, the reconstruction of training and out-of-sample data is Eq. (\ref{equation_reconstruction_dual_SPCA}) and (\ref{equation_outOfSample_reconstruction_dual_SPCA}), respectively. 

Note that in RDA and dual RDA, we can truncate $\b{U}$, $\b{\Sigma}$, and $\b{V}$, respectively, to $\b{U} \in \mathbb{R}^{d \times p}$, $\b{\Sigma} \in \mathbb{R}^{p \times p}$, and $\b{V} \in \mathbb{R}^{d \times p}$ in order to have a $p$-dimensional subspace ($p \leq d$). 
For determining the appropriate $p$ in RDA, the scree plot \cite{cattell1966scree} or the ratio $\lambda_j / \sum_{k=1}^d \lambda_k$ \cite{abdi2010principal} can be used, where $\lambda_j$ is the $j$-th largest eigenvalue. 

The dual RDA is very useful especially if the dimensionality of data is much greater than the sample size of data, i.e., $d \gg n$.
In this case $\b{W} \in \mathbb{R}^{d \times k}=\mathbb{R}^{d \times n}$. 
According to Eq. (\ref{equation_W_SVD}), $\b{U}$ and $\b{V}$ are the eigenvectors of $\b{W}\b{W}^\top \in \mathbb{R}^{d \times n}$ and $\b{W}^\top\b{W} \in \mathbb{R}^{n \times n}$, respectively. As $n \ll d$ in this case, the computation of eigenvectors of $\b{W}^\top\b{W}$ is much faster and needs less storage than $\b{W}\b{W}^\top$. Therefore, in this case, we use the eigenvalue decomposition:
\begin{align}
\b{W}^\top\b{W} = \b{V} \b{\Sigma}^2 \b{V}^\top,
\end{align}
rather than using Eq. (\ref{equation_W_SVD}). Note that in dual RDA, we do not require $\b{U}$ but only $\b{\Sigma}$ and $\b{V}$. Hence, if $n \ll d$, it is recommended to use dual RDA which is more efficient than RDA in terms of speed of calculation and storage. 

\section{Kernel RDA}\label{section_kernel_RDA}

Let $\b{\phi}: \mathcal{X} \rightarrow \mathcal{H}$ be the pulling function mapping the data $\b{x} \in \mathcal{X}$ to the feature space $\mathcal{H}$. In other words, $\b{x} \mapsto \b{\phi}(\b{x})$. Let $t$ denote the dimensionality of the feature space, i.e., $\b{\phi}(\b{x}) \in \mathbb{R}^t$ while $\b{x} \in \mathbb{R}^d$. Note that we usually have $t \gg d$.
The kernel over two vectors $\b{x}_1$ and $\b{x}_2$ is the inner product of their pulled data \cite{hofmann2008kernel}:
\begin{align}\label{equation_kernel_scalar}
\mathbb{R} \ni \b{k}(\b{x}_1, \b{x}_2) := \b{\phi}(\b{x}_1)^\top \b{\phi}(\b{x}_2).
\end{align}
We can have two types of kernel RDA, one using the dual RDA and kernel trick and the other one using the representation theory \cite{alperin1993local}. The former, which makes use of the inner product of data instances, holds for only two special cases but the latter works for the entire range of the Roweis map. 

\subsection{Kernel RDA Using Dual RDA for Special Cases}

If $r_2=0$, we have the dual RDA and can use it for working out the kernel RDA using the kernel trick. However, the dual RDA is useful for the kernel trick in the two cases of $r_1=0$ and $r_1=1$. The reason is that in these cases, the inner product of data points appear enabling us to use the kernel trick. Therefore, we can use the dual RDA for kernel trick in $r_1=r_2=0$ (i.e., PCA) and $r_1=0, r_2=1$ (i.e., SPCA). 
This explains why, in the literature, PCA and SPCA (with $r_2=0$) have their kernel methods using kernel trick (or their dual) but kernel FDA (with $r_1=0, r_2=1$) uses representation theory and not the kernel trick.

\subsubsection{Special Case of Kernel RDA: Kernel PCA}

If $r_1=0$, we have $\b{R}_1 = \b{S}_T = \breve{\b{X}} \breve{\b{X}}^\top$ according to Eq. (\ref{equation_R1_decompose_1}). In this case, $\b{W}=\breve{\b{X}}=\b{X}\b{H}$. So, we decompose $\breve{\b{X}}$ in Eq. (\ref{equation_W_SVD}).
In this case, RDA is reduced to PCA where the data should be centered. Hence, in reconstruction, we add the training mean back. 
Therefore, the Eqs. (\ref{equation_projected_dual_RDA}), (\ref{equation_outOfSample_projected_dual_RDA}), (\ref{equation_reconstruction_dual_SPCA}), and (\ref{equation_outOfSample_reconstruction_dual_SPCA}) are modified to:
\begin{align}
&\breve{\b{X}} = \b{U}\b{\Sigma}\b{V}^\top \overset{(a)}{\implies} \b{U}^\top \breve{\b{X}} = \b{\Sigma} \b{V}^\top \overset{(\ref{equation_projection_severalPoint})}{\implies} \widetilde{\b{X}} = \b{\Sigma} \b{V}^\top, \label{equation_dual_PCA_projection} \\
&\widetilde{\b{X}}_t = \b{\Sigma}^{-1}\b{V}^\top \breve{\b{X}}^\top\, \breve{\b{X}}_t, \label{equation_dual_PCA_outOfSample_projection} \\
&\breve{\b{X}} = \b{U}\b{\Sigma}\b{V}^\top \overset{(a)}{\implies} \b{U} = \breve{\b{X}} \b{V} \b{\Sigma}^{-1} \nonumber \\
&~~~~~~~~~~~ \implies \widehat{\b{X}} \overset{(\ref{equation_reconstruction_severalPoint})}{=} \b{U}\widetilde{\b{X}} + \b{\mu} \overset{(\ref{equation_dual_PCA_projection})}{=} \breve{\b{X}} \b{V} \b{V}^\top + \b{\mu}, \label{equation_dual_PCA_reconstruction} \\
&\widehat{\b{X}}_t = \breve{\b{X}}\b{V}\b{\Sigma}^{-2}\b{V}^\top\breve{\b{X}}^\top \breve{\b{X}}_t + \b{\mu}, \label{equation_dual_PCA_outOfSample_reconstruction}
\end{align}
respectively, where $(a)$ is because $\b{U}$ and $\b{V}$ are orthogonal matrices and $\breve{\b{X}}_t$ is the centered out-of-sample data using the training mean:
\begin{align}
\breve{\b{X}}_t := \b{X}_t - \b{\mu}.
\end{align}

In the kernel RDA for $r_1=r_2=0$ (i.e., kernel PCA \cite{scholkopf1997kernel,scholkopf1998nonlinear}), the incomplete SVD is applied on the $\breve{\b{\Phi}}(\b{X}) := \b{\Phi}(\b{X}) \b{H}$ which is the centered data in the feature space.
We use the Eqs. (\ref{equation_dual_PCA_projection}), (\ref{equation_dual_PCA_outOfSample_projection}), (\ref{equation_dual_PCA_reconstruction}), and (\ref{equation_dual_PCA_outOfSample_reconstruction}) and replace the inner products by the kernels:
\begin{align}
&\breve{\b{\Phi}}(\b{X}) = \b{U}\b{\Sigma}\b{V}^\top \implies \b{\Phi}(\widetilde{\b{X}}) = \b{\Sigma} \b{V}^\top, \label{equation_kernel_PCA_projection} \\
&\b{\Phi}(\widetilde{\b{X}}_t) = \b{\Sigma}^{-1}\b{V}^\top \breve{\b{\Phi}}(\b{X})^\top\, \breve{\b{\Phi}}(\b{X}_t) \overset{(a)}{=} \b{\Sigma}^{-1}\b{V}^\top \breve{\b{K}}_t, \label{equation_kernel_PCA_outOfSample_projection} \\
&\b{\Phi}(\widehat{\b{X}}) = \breve{\b{\Phi}}(\b{X}) \b{V} \b{V}^\top + \b{\mu}, \label{equation_kernel_PCA_reconstruction} \\
&\b{\Phi}(\widehat{\b{X}}_t) = \breve{\b{\Phi}}(\b{X})\b{V}\b{\Sigma}^{-2}\b{V}^\top\breve{\b{\Phi}}(\b{X})^\top \breve{\b{\Phi}}(\b{X}_t) + \b{\mu} \nonumber \\
&~~~~~~~~~~~\overset{(a)}{=} \breve{\b{\Phi}}(\b{X})\b{V}\b{\Sigma}^{-2}\b{V}^\top \breve{\b{K}}_t + \b{\mu}, \label{equation_kernel_PCA_outOfSample_reconstruction}
\end{align}
where $\breve{\b{\Phi}}(\b{X}_t)$ is the centered out-of-sample data in the feature space using the mean of the pulled training data and $(a)$ is because the double-centered kernel over the training and out-of-sample data is calculated as:
\begin{align}
\mathbb{R}^{n \times n_t} \ni \breve{\b{K}}_t &:= \b{K}_t - \frac{1}{n} \b{1}_{n \times n}\, \b{K}_t - \frac{1}{n} \b{K}_x\, \b{1}_{n \times n_t} \nonumber \\
&~~~~ + \frac{1}{n^2} \b{1}_{n \times n}\, \b{K}_x\, \b{1}_{n \times n_t}, \label{equation_doubleCentered_outOfSample_kernel}
\end{align}
where $\mathbb{R}^{n \times n_t} \ni \b{K}_t := \b{\Phi}(\b{X})^\top \b{\Phi}(\b{X}_t)$ and $\mathbb{R}^{n \times n} \ni \b{K}_x := \b{\Phi}(\b{X})^\top \b{\Phi}(\b{X})$.
For the proof of Eq. (\ref{equation_doubleCentered_outOfSample_kernel}), refer to the appendices in \cite{scholkopf1998nonlinear} and \cite{ghojogh2019unsupervised}.
In Eqs. (\ref{equation_kernel_PCA_reconstruction}) and (\ref{equation_kernel_PCA_outOfSample_reconstruction}), there exists $\breve{\b{\Phi}}(\b{X})$ which is not necessarily available; therefore, in kernel RDA with $r_1=r_2=0$ (kernel PCA), we cannot reconstruct. 

In practice, the $\breve{\b{\Phi}}(\b{X})$ is not available for its SVD decomposition; therefore, the $\b{V}$ and $\b{\Sigma}$ are found by eigenvalue problem for the double-centered training kernel, $\breve{\b{K}}_x := \b{H} \b{K}_x \b{H} = \breve{\b{\Phi}}(\b{X})^\top \breve{\b{\Phi}}(\b{X})$:
\begin{align}
\breve{\b{K}}_x \b{V} = \b{V} \b{\Sigma}^2.
\end{align}

\subsubsection{Special Case of Kernel RDA: Kernel SPCA}

If $r_1=1$, we have $\b{R}_1 = \b{Q} \b{Q}^\top$ according to Eq. (\ref{equation_R1_decompose_1}). In this case, $\b{W}=\b{Q}=\b{X} \b{H} \b{\Upsilon}$. So, we decompose $\b{Q}$ in Eq. (\ref{equation_W_SVD}).
In this case, RDA is reduced to SPCA. 
Therefore, the Eqs. (\ref{equation_projected_dual_RDA}), (\ref{equation_outOfSample_projected_dual_RDA}), (\ref{equation_reconstruction_dual_SPCA}), and (\ref{equation_outOfSample_reconstruction_dual_SPCA}) are modified to:
\begin{align}
&\widetilde{\b{X}} = \b{\Sigma}^{-1} \b{V}^\top \b{\Upsilon}^\top \b{H} \b{X}^\top \b{X}, \label{equation_dual_SPCA_projection} \\
&\widetilde{\b{X}}_t = \b{\Sigma}^{-1} \b{V}^\top \b{\Upsilon}^\top \b{H} \b{X}^\top \b{X}_t, \label{equation_dual_SPCA_outOfSample_projection} \\
&\widehat{\b{X}} = \b{X} \b{H} \b{\Upsilon} \b{V} \b{\Sigma}^{-2} \b{V}^\top \b{\Upsilon}^\top \b{H} \b{X}^\top \b{X}, \label{equation_dual_SPCA_reconstruction} \\
&\widehat{\b{X}}_t = \b{X} \b{H} \b{\Upsilon} \b{V} \b{\Sigma}^{-2} \b{V}^\top \b{\Upsilon}^\top \b{H} \b{X}^\top \b{X}_t. \label{equation_dual_SPCA_outOfSample_reconstruction}
\end{align}
In the kernel RDA for $r_1=1, r_2=0$ (i.e., kernel SPCA \cite{barshan2011supervised}), the incomplete SVD is applied on the $\breve{\b{\Phi}}(\b{X})$.
We replace the inner products by the kernels in the Eqs. (\ref{equation_dual_SPCA_projection}), (\ref{equation_dual_SPCA_outOfSample_projection}), (\ref{equation_dual_SPCA_reconstruction}), and (\ref{equation_dual_SPCA_outOfSample_reconstruction}):
\begin{align}
&\b{\Phi}(\widetilde{\b{X}}) = \b{\Sigma}^{-1} \b{V}^\top \b{\Upsilon}^\top \b{H} \b{K}_x, \label{equation_kernel_SPCA_projection} \\
&\b{\Phi}(\widetilde{\b{X}}_t) = \b{\Sigma}^{-1} \b{V}^\top \b{\Upsilon}^\top \b{H} \b{K}_t, \label{equation_kernel_SPCA_outOfSample_projection} \\
&\b{\Phi}(\widehat{\b{X}}) = \b{\Phi}(\b{X}) \b{H} \b{\Upsilon} \b{V} \b{\Sigma}^{-2} \b{V}^\top \b{\Upsilon}^\top \b{H} \b{K}_x, \label{equation_kernel_SPCA_reconstruction} \\
&\b{\Phi}(\widehat{\b{X}}_t) = \b{\Phi}(\b{X}) \b{H} \b{\Upsilon} \b{V} \b{\Sigma}^{-2} \b{V}^\top \b{\Upsilon}^\top \b{H} \b{K}_t. \label{equation_kernel_SPCA_outOfSample_reconstruction}
\end{align}
In Eqs. (\ref{equation_kernel_SPCA_reconstruction}) and (\ref{equation_kernel_SPCA_outOfSample_reconstruction}), the term $\b{\Phi}(\b{X})$ exists which is not necessarily available so reconstruction cannot be done in kernel RDA with $r_1=1, r_2=0$ (i.e., kernel SPCA).
Note that, in practice, the $\b{V}$ and $\b{\Sigma}$ are found by the eigenvalue problem $\b{K}_x \b{V} = \b{V} \b{\Sigma}^2$.

\subsection{Direct Kernel RDA}

\subsubsection{Methodology}

According to the representation theory \cite{alperin1993local}, any pulled solution (direction) $\b{\phi}(\b{u}) \in \mathcal{H}$ must lie in the span of all the training vectors pulled to $\mathcal{H}$, i.e., $\b{\Phi}(\b{X}) = [\b{\phi}(\b{x}_1), \dots, \b{\phi}(\b{x}_n)] \in \mathbb{R}^{t\times n}$. Hence:
\begin{align}\label{equation_phi_u}
\mathbb{R}^{t} \ni \b{\phi}(\b{u}) = \sum_{i=1}^n \theta_i\, \b{\phi}(\b{x}_i) = \b{\Phi}(\b{X})\, \b{\theta},
\end{align}
where $\mathbb{R}^n \ni \b{\theta} = [\theta_1, \dots, \theta_n]^\top$ is the unknown vector of coefficients, and $\b{\phi}(\b{u}) \in \mathbb{R}^t$ is the pulled RDA direction to the feature space.
The pulled directions can be put together in $\mathbb{R}^{t \times p} \ni \b{\Phi}(\b{U}) = [\b{\phi}(\b{u}_1), \dots, \b{\phi}(\b{u}_p)]$:
\begin{align}\label{equation_Phi_U}
\mathbb{R}^{t \times p} \ni \b{\Phi}(\b{U}) = \b{\Phi}(\b{X})\, \b{\Theta},
\end{align}
where $\mathbb{R}^{n \times p} \ni \b{\Theta} = [\b{\theta}_1, \dots, \b{\theta}_p]$.

In order to have RDA in the feature space, we first kernelize the objective function of the Eq. (\ref{equation_optimization_RDA}):
\begin{align}
&\textbf{tr}\big(\b{\Phi}(\b{U})^\top \b{\Phi}(\b{R}_1)\, \b{\Phi}(\b{U})\big) \nonumber \\ 
&\overset{(\ref{equation_R1})}{=} \textbf{tr}\big(\b{\Phi}(\b{U})^\top \b{\Phi}(\b{X}) \b{H} \b{P} \b{H} \b{\Phi}(\b{X})^\top \b{\Phi}(\b{U})\big) \nonumber \\
&\overset{(\ref{equation_Phi_U})}{=} \textbf{tr}\big(\b{\Theta}^\top \b{\Phi}(\b{X})^\top \b{\Phi}(\b{X}) \b{H} \b{P} \b{H} \b{\Phi}(\b{X})^\top \b{\Phi}(\b{X})\, \b{\Theta}\big) \nonumber \\
&\overset{(a)}{=} \textbf{tr}\big(\b{\Theta}^\top \b{K}_x\, \b{H} \b{P} \b{H} \b{K}_x\, \b{\Theta}\big) \overset{(b)}{=} \textbf{tr}\big(\b{\Theta}^\top \b{M}\, \b{\Theta}\big), \label{equation_theta_M_theta}
\end{align}
where $(a)$ is because the kernel matrix over $\b{X}$ is defined as:
\begin{align}\label{equation_kernel_over_X}
\mathbb{R}^{n \times n} \ni \b{K}_x := \b{\Phi}(\b{X})^\top \b{\Phi}(\b{X}),
\end{align}
and $(b)$ is because:
\begin{align}\label{equation_M}
\mathbb{R}^{n \times n} \ni \b{M} := \b{K}_x\, \b{H} \b{P} \b{H} \b{K}_x.
\end{align}

In order to kernelize the constraint in the Eq. (\ref{equation_optimization_RDA}), it is easier to first consider a one-dimensional subspace and then extend it to multi-dimensional subspace. We can prove that:
\begin{align}
\b{\phi}(\b{u})^\top \b{\Phi}(\b{S}_W)\, \b{\phi}(\b{u}) &= \b{\theta}^\top \Big( \sum_{j=1}^c \b{K}_j \b{H}_j \b{K}_j^\top \Big) \b{\theta} \overset{(a)}{=} \b{\theta}^\top \b{N}\, \b{\theta}, \label{equation_d_W_pulled}
\end{align}
where $c$ is the number of classes, $n_j$ is the sample size of the $j$-th class, and:
\begin{align}
&\mathbb{R}^{n_j \times n_j} \ni \b{H}_j := \b{I} - (1/n_j) \b{1}\b{1}^\top, \label{equation_H_j} \\
&\mathbb{R}^{n \times n_j} \ni \b{K}_j := \b{\Phi}(\b{X})^\top \b{\Phi}(\b{X}_j), \label{equation_K_j}
\end{align}
and $(a)$ is because:
\begin{align}\label{equation_N}
\mathbb{R}^{n \times n} \ni \b{N} := \sum_{j=1}^c \b{K}_j \b{H}_j \b{K}_j^\top.
\end{align}
The derivation of Eq. (\ref{equation_d_W_pulled}) is in Appendix A.

If the subspace is one-dimensional, the constraint in the Eq. (\ref{equation_optimization_RDA}) is kernelized as:
\begin{align*}
&\b{\phi}(\b{u})^\top \b{\Phi}(\b{R}_2)\, \b{\phi}(\b{u}) \\
&\overset{(\ref{equation_R2})}{=} r_2\, \b{\phi}(\b{u})^\top \b{\Phi}(\b{S}_W)\, \b{\phi}(\b{u}) + (1 - r_2)\, \b{\phi}(\b{u})^\top \b{\phi}(\b{u}) \\
&\overset{(a)}{=} r_2\, \b{\theta}^\top \b{N} \b{\theta} + (1 - r_2)\, \b{\theta}^\top \b{K}_x\, \b{\theta} \\
&= \b{\theta}^\top \big( r_2\, \b{N} + (1-r_2)\, \b{K}_x \big)\, \b{\theta} \overset{(b)}{=} \b{\theta}^\top \b{L}\, \b{\theta},
\end{align*}
where $(a)$ is because of Eqs. (\ref{equation_phi_u}), (\ref{equation_kernel_over_X}), and (\ref{equation_d_W_pulled}), and $(b)$ is because:
\begin{align}\label{equation_L}
\mathbb{R}^{n \times n} \ni \b{L} := r_2\, \b{N} + (1-r_2)\, \b{K}_x.
\end{align}
Similarly, we can extend to multi-dimensional subspace:
\begin{align}
\textbf{tr}\big(\b{\phi}(\b{U})^\top \b{\Phi}(\b{R}_2)\, \b{\phi}(\b{U})\big) = \textbf{tr}(\b{\Theta}^\top \b{L}\, \b{\Theta}). \label{equation_theta_L_theta}
\end{align}
According to Eqs. (\ref{equation_optimization_RDA_criterion}), (\ref{equation_theta_M_theta}), and (\ref{equation_theta_L_theta}), the Roweis criterion in the feature space is:
\begin{align}\label{equation_RDA_criterion_in_feature_space}
f(\b{\Theta}) := \frac{\textbf{tr}(\b{\Theta}^\top \b{M}\, \b{\Theta})}{\textbf{tr}(\b{\Theta}^\top \b{L}\, \b{\Theta})}.
\end{align}
Maximizing this generalized Rayleigh-Ritz Quotient \cite{parlett1998symmetric} is equivalent to:
\begin{equation}\label{equation_optimization_direct_kernel_RDA}
\begin{aligned}
& \underset{\b{\Theta}}{\text{maximize}}
& & \textbf{tr}(\b{\Theta}^\top \b{M}\, \b{\Theta}), \\
& \text{subject to}
& & \b{\Theta}^\top \b{L}\, \b{\Theta} = \b{I},
\end{aligned}
\end{equation}
whose solution is the generalized eigenvalue problem $(\b{M}, \b{L})$ according to Eq. (\ref{equation_scatter_generalized_eigendecomposition}). Therefore, the direct kernel RDA directions are the eigenvectors of this generalized eigenvalue problem. The directions are sorted from the leading to trailing eigenvalues because of the maximization in Eq. (\ref{equation_optimization_direct_kernel_RDA}). 
In kernel RDA, the directions are $n$-dimensional while in RDA, we had $d$-dimensional directions.

In kernel RDA, the projection and reconstruction of the training and out-of-sample data are:
\begin{align}
&\widetilde{\b{X}} = \b{\Phi}(\b{U})^\top \b{\Phi}(\b{X}) \overset{(\ref{equation_Phi_U})}{=} \b{\Theta}^\top \b{K}_x, \\
&\widehat{\b{X}} = \b{\Phi}(\b{U}) \b{\Phi}(\b{U})^\top \b{\Phi}(\b{X}) \overset{(\ref{equation_Phi_U})}{=} \b{\Phi}(\b{X}) \b{\Theta} \b{\Theta}^\top \b{K}_x, \\
&\widetilde{\b{X}}_t = \b{\Theta}^\top \b{K}_t, ~~~ \widehat{\b{X}}_t = \b{\Phi}(\b{X}) \b{\Theta} \b{\Theta}^\top \b{K}_t,
\end{align}
where $\b{\Phi}(\b{X})$ existing in the reconstructions are not necessarily available so we do not have reconstruction in kernel RDA.

\subsubsection{Properties of $\b{M}$, $\b{N}$, and $\b{L}$}

According to Eqs. (\ref{equation_M}), (\ref{equation_L}), and (\ref{equation_P}), we have:
\begin{align}
\b{M} &= 
\left\{
    \begin{array}{ll}
        \b{K}_x\, \b{H} \b{K}_y\, \b{H} \b{K}_x & \text{if } r_1=1, \\
        \b{K}_x\, \b{H} \b{K}_x & \text{if } r_1=0,
    \end{array}
\right. \label{equation_M_bracket} \\
\b{L} &= 
\left\{
    \begin{array}{ll}
        \b{N} \overset{(\ref{equation_N})}{=} \sum_{j=1}^c \b{K}_j \b{H}_j \b{K}_j^\top & \text{if } r_2=1, \\
        \b{K}_x & \text{if } r_2=0,
    \end{array}
\right. \label{equation_L_bracket}
\end{align}
where it is noticed that $\b{H}$ is idempotent. 
We see that if $r_1=1$ and $r_2=1$, the labels are used in calculation of $\b{M}$ and $\b{N}$, respectively. Also, if if $r_1=0$ and $r_2=0$, the labels are not used in them at all. Thus, the Roweis factors are measures of using labels also in kernel RDA. The Eq. (\ref{equation_s}) can again be used as level of supervision.  
Moreover, according to Eqs. (\ref{equation_M_bracket}), (\ref{equation_L_bracket}), and (\ref{equation_N}), the matrices $\b{M}$, $\b{N}$, and $\b{L}$ are symmetric.

About the ranks of these matrices, we have $\textbf{rank}(\b{P}) \leq \min(n,t) + n$ according to Eq. (\ref{equation_rank_P}).
As $\mathbb{R}^{n \times n} \ni \b{K}_x = \b{\Phi}(\b{X})^\top \b{\Phi}(\b{X})$, we have $\textbf{rank}(\b{K}_x) \leq \min(n,t)$.
According to Eq. (\ref{equation_M}):
\begin{align}
\textbf{rank}(\b{M}) &\leq \min\big( \textbf{rank}(\b{K}_x) + \textbf{rank}(\b{H}) + \textbf{rank}(\b{P}) \big) \nonumber \\
&\leq \min(n,n-1,t) \overset{(a)}{=} n-1, \label{equation_rank_M}
\end{align}
where the $-1$ is because of the centering the matrix and $(a)$ is because $t$ is often large.
In both the extreme cases $r_1=0$ and $r_1=1$, the rank of $\b{M}$ is again at most $n-1$.
According to Eq. (\ref{equation_N}), we have $\textbf{rank}(\b{N}) \leq \min(n,c-1)$ because it has $c$ iterations each of which includes the centring matrix.
According to Eq. (\ref{equation_L}) and the subadditivity property of the rank:
\begin{align}\label{equation_rank_L}
\textbf{rank}(\b{L}) \leq \textbf{rank}(\b{N}) + \textbf{rank}(\b{K}_x) \leq \min(n,c)-1.
\end{align}
In the extreme cases $r_2=0$ and $r_2=1$, the rank of $\b{L}$ is at most $\min(n,t)=n$ and $\min(n,c-1)$, respectively. 

\subsubsection{Dimensionality of the Direct Kernel RDA Subspace}

Similar to Eq. (\ref{equation_RDA_dirty_solution}), we cal solve Eq. (\ref{equation_optimization_direct_kernel_RDA}) as $\b{\Theta} = \textbf{eig}(\b{L}^{-1} \b{M})$ where the numeric hack of Eq. (\ref{equation_RDA_dirty_solution_singular}) may also be required. 
According to Eqs. (\ref{equation_rank_M}) and (\ref{equation_rank_L}), we have:
\begin{align}
\textbf{rank}(\b{L}^{-1} \b{M}) &\leq  \min\big(\textbf{rank}(\b{L}^{-1}), \textbf{rank}(\b{M})\big) \nonumber \\
&\leq \min(n,c) - 1. \label{equation_rank_kernel_RDA}
\end{align}
Therefore, the dimensionality of the kernel RDA subspace is $p \leq \min(n,c) - 1$, restricted by rank of $\b{L}$, where the $\min(n-1,c-1)$ leading eigenvectors are considered as the kernel RDA directions and the rest of eigenvectors are invalid with very small eigenvalues. Notice that in most datasets, $c \ll n$ so we usually have $p \leq c-1$. Recall that this upperbound also exists on the dimensionality of the FDA and kernel FDA subspaces \cite{friedman2001elements,mika1999fisher,ghojogh2019fisher}.

\subsubsection{Special Cases of Kernel RDA}

The Roweis map can have two layers, one for the input space and another for the feature space. The top layer is the bottom layer pulled to the feature space (see Fig. \ref{figure_Roweis_map}-b). Therefore, the four corners of Roweis map on the feature space can be considered as kernel PCA, kernel FDA, kernel FDA, and kernel DSDA. The whole map includes the kernel methods of an infinite number of subspace learning algorithms. 
Kernel PCA \cite{scholkopf1997kernel,scholkopf1998nonlinear}, kernel FDA \cite{mika1999fisher,mika2000invariant}, and kernel SPCA \cite{barshan2011supervised} are already proposed in the literature but the kernel method of the other algorithms in the Rowies map, such as kernel DSDA, are new. 

As mentioned before, the methods in the Roweis map along $r_2=0$, including PCA and SPCA, have two kinds of kernelization which are using kernel trick (or dual of each method) and representation theory. This explains why there exist two types of kernel SPCA \cite{barshan2011supervised}. The kernel PCA using the kernel trick is already proposed \cite{scholkopf1997kernel,scholkopf1998nonlinear} while the kernel PCA using representation theory is proposed here for the first time to the best of our knowledge.
The direct kernel RDA is reduced to exactly the existing direct kernel SPCA \cite{barshan2011supervised} for $r_1=1, r_2=0$.



\section{Experiments}\label{section_experiments}

\begin{figure*}[!t]
\centering
\includegraphics[width=6.5in]{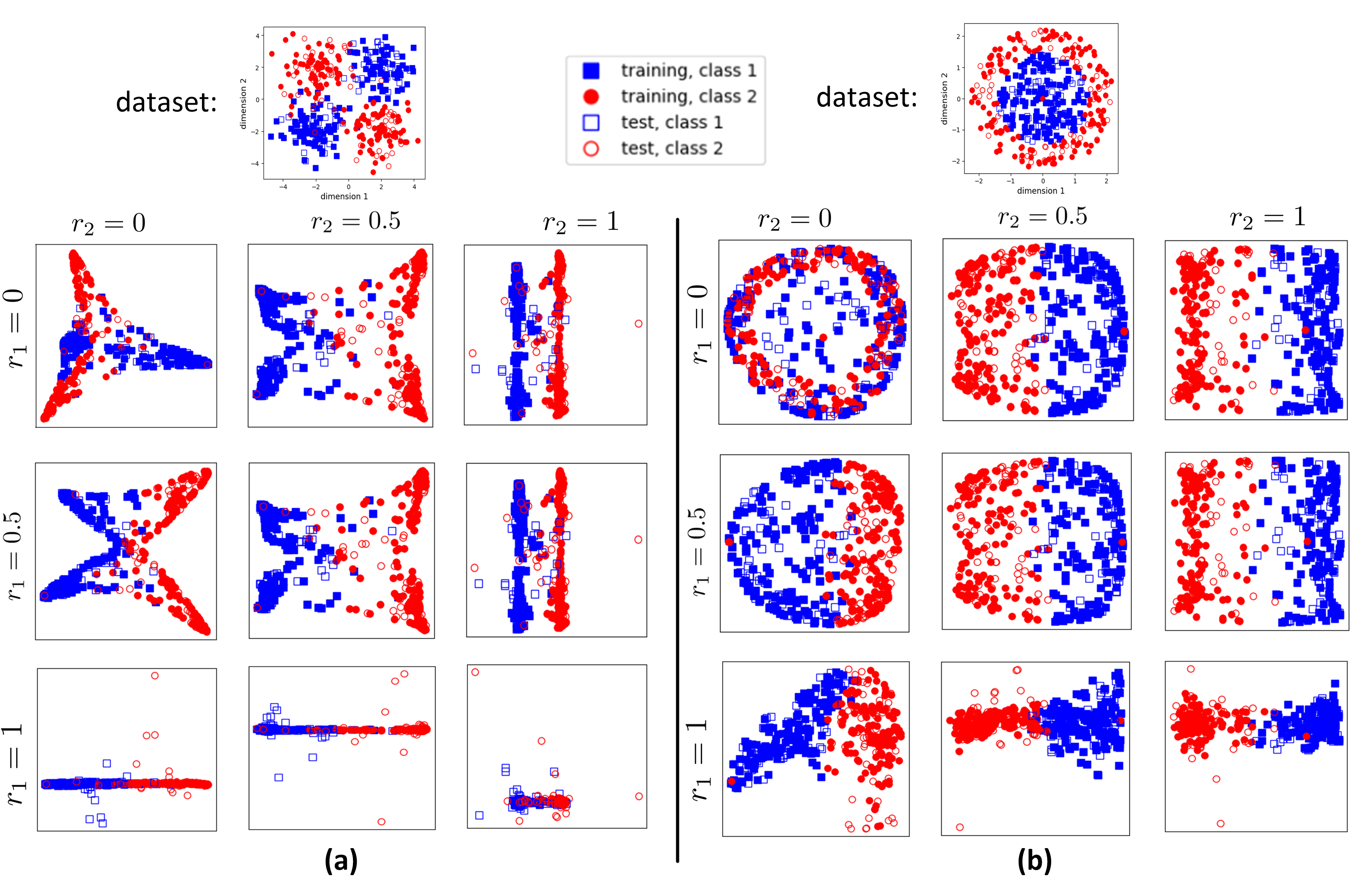}
\caption{The first two dimensions of the projected data in kernel RDA for (a) binary XOR dataset and (b) concentric rings dataset.}
\label{figure_synthetic_projections}
\end{figure*}

\subsection{Visualization: Synthetic Nonlinear Datasets}

One of the applications of subspace learning is data visualization. For the first experiment, we created two synthetic datasets which are nonlinear. The two datasets are binary XOR and concentric rings having two dimensions and two classes. 
The sample size in every dataset is $400$ where $70\%$ and $30\%$ of the data are used for training and out-of-sample (test), respectively. 
The datasets are shown in Fig. \ref{figure_synthetic_projections}. 

As these datasets are highly nonlinear, RDA does not separate the classes as well as kernel RDA. Therefore, the first two dimensions of the embedding in nine special cases of kernel RDA (with Radial Basis Function (RBF) kernel for $\b{K}_x$) are illustrated in Fig. \ref{figure_synthetic_projections}. 
Note that in all the classification experiments in this paper, we used the Kronecker delta kernel for $\b{K}_y$ \cite{barshan2011supervised}.
As this figure shows, kernel PCA does not perform as well as kernel SPCA, kernel FDA, and kernel DSDA. Overall, the larger the Roweis factors get, the better the two classes are separated which is expected because the supervision level is increased. By sweeping $r_2=0 \rightarrow 1$, the two classes are almost collapsed into two one dimensional lines because, according to Eq. (\ref{equation_rank_kernel_RDA}), the rank of kernel RDA is restricted by $c-1=1$ when $r_2=1$ but this restriction does not exist for $r_2=0$. 
another interpretation is because of taking the within scatter into account when $r_2$ is closer to one so the classes are collapsed. 
Moreover, these figures show that the RDA is capable of handling out-of-sample data well enough.

\subsection{Rowiesfaces: The Ghost Faces in RDA}

For the next experiment, we used the AT\&T face dataset \cite{web_ATT_face_dataset} including $40$ subjects each having $10$ images with different expressions and poses. 
The data were standardized to have zero mean and unit variance.
We divided the dataset into two classes of images having and not having eye glasses. 

\subsubsection{The Facial Eigenvectors}

We trained nine special cases, $r_1, r_2 \in \{0, 0.5, 1\}$, of RDA and kernel RDA using the facial dataset. 
We name the facial eigenvectors, or ghost faces, in RDA as \textit{Roweisfaces}. The existing special cases of Roweisfaces in the literature are eigenfaces ($r_1=r_2=0$) \cite{turk1991eigenfaces,turk1991face}, Fisherfaces ($r_1=0, r_2=1$) \cite{belhumeur1997eigenfaces}, and supervised eigenfaces ($r_1=1, r_2=0$) \cite{barshan2011supervised,ghojogh2019unsupervised}. 
For $r_1=r_2=1$ in Roweisfaces, we use the name \textit{double supervised eigenfaces}. 
We name facial embedding using kernel RDA as \textit{kernel Roweisfaces} whose existing special cases are kernel eigenfaces ($r_1=r_2=0$) \cite{yang2000face}, kernel Fisherfaces ($r_1=0, r_2=1$) \cite{yang2002kernel}, kernel supervised eigenfaces ($r_1=1, r_2=0$) \cite{barshan2011supervised,ghojogh2019unsupervised}. For $r_1=r_2=1$ in kernel Roweisfaces, we use the name \textit{kernel double supervised eigenfaces}. 

\begin{figure*}[!t]
\centering
\includegraphics[width=5.5in]{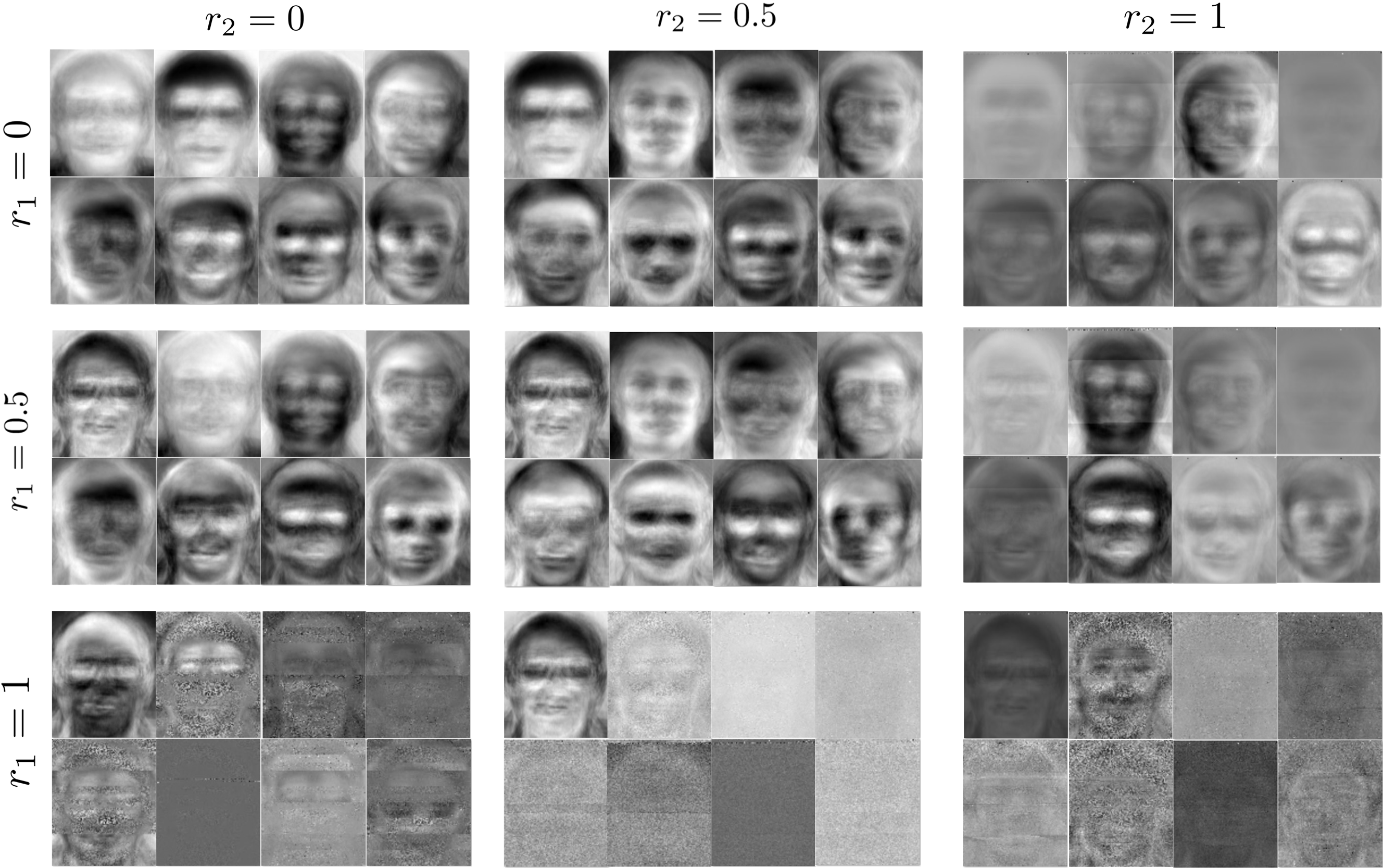}
\caption{The Roweisfaces: the eight leading eigenvectors for the special cases in the Roweis map where the first, fifth, and eighth eigenvectors of every case are at the top-left, bottom-left, and bottom-right, respectively.}
\label{figure_Roweisfaces}
\end{figure*}

The eigenvectors in kernel RDA are $n$-dimensional and not $d$-dimensional so we can show the ghost faces only in Roweisfaces and not kernel Roweisfaces. The trained Roweisfaces are shown in Fig. \ref{figure_Roweisfaces} for the special cases. The PCA case has captured different features such as eyes, hair, lips, nose, and face border. However, the more we consider the labels by increasing $r_1$ and $r_2$, the more features related to eyes and cheeks are extracted because of the more discrimination of having or not having glasses. Increasing $r_1$ tends to extract more features like Haar wavelet features \cite{stankovic2003haar} which are useful for face feature detection (see Viola-Jones face detector \cite{wang2014analysis}). Increasing $r_2$, however, fades out the irrelevant features leaving merely the eyes which are important. The double supervised eigenfaces have a mixture of Haar features and fading out unimportant features.

\subsubsection{The Projections}

\begin{figure*}[!t]
\centering
\includegraphics[width=6.5in]{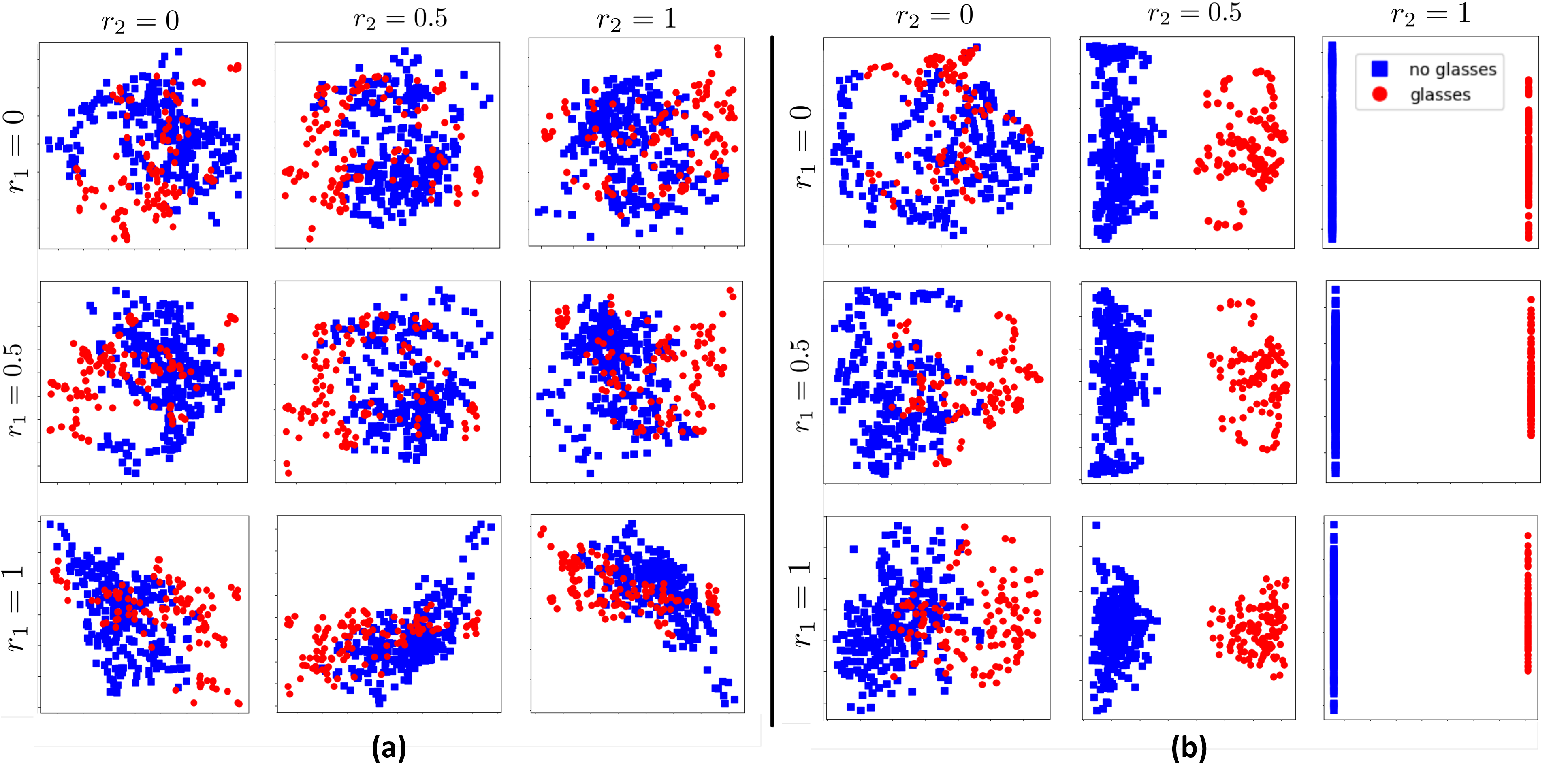}
\caption{The first two dimensions of the projected data in (a) Roweisfaces and (b) kernel Roweisfaces.}
\label{figure_Roweisfaces_scatters}
\end{figure*}

The top two dimensions of projection of the facial images into the RDA and kernel RDA (with RBF kernel for $\b{K}_x$) subspaces are shown in Fig. \ref{figure_Roweisfaces_scatters}.
As expected, kernel RDA separates the classes better that RDA. Also, the larger the supervision level, the better the separation. For the same reasons explained for Fig. \ref{figure_synthetic_projections}, the two classes are collapsed into one-dimensional lines for $r_2=1$ in kernel RDA. 

\subsubsection{The Reconstructions}

\begin{figure}[!t]
\centering
\includegraphics[width=3.3in]{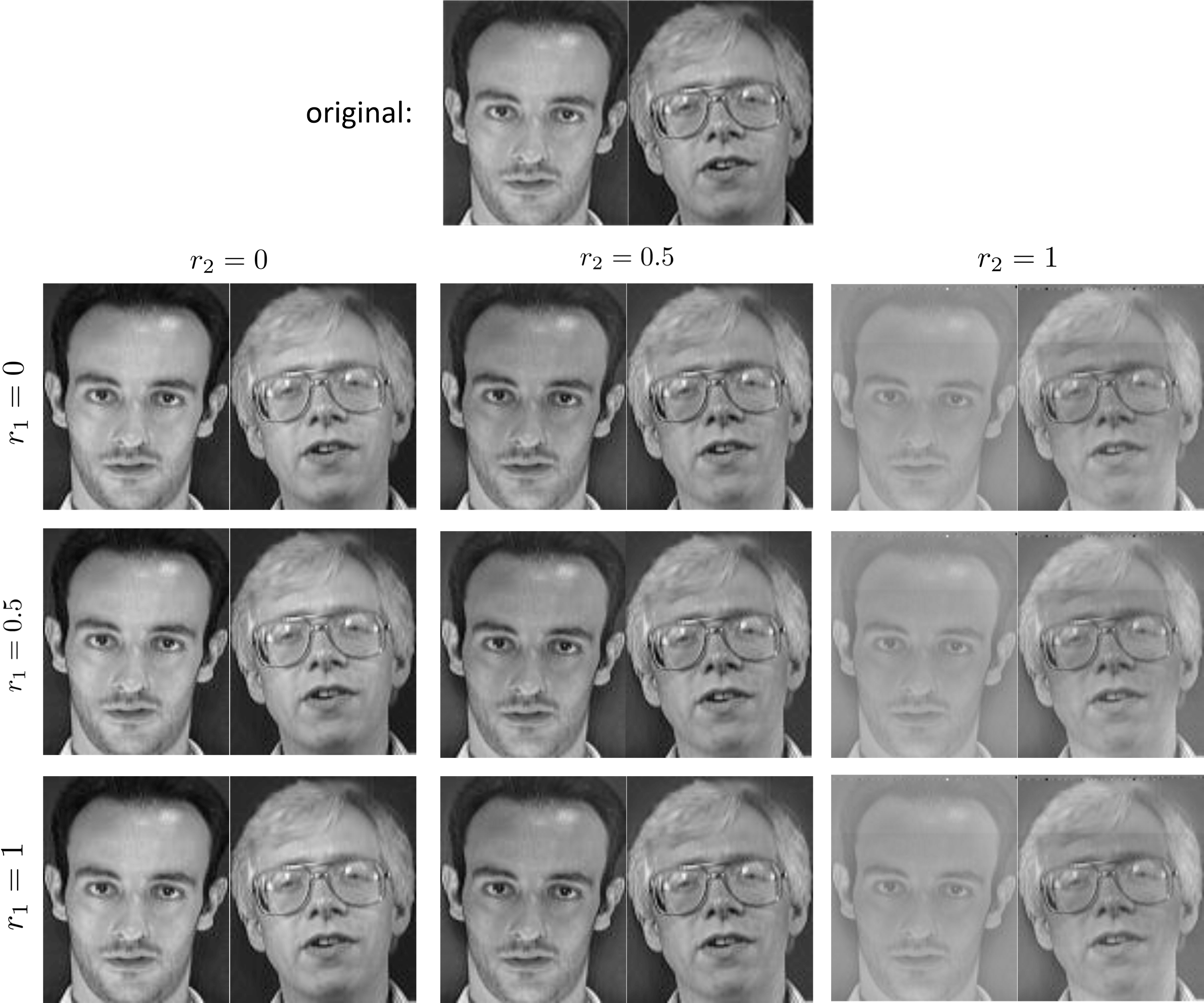}
\caption{The reconstructed images after projection into the RDA subspaces.}
\label{figure_reconstructed}
\end{figure}

As explained before, the data cannot be reconstructed in kernel RDA but it can be done in RDA. Some reconstructed images for the facial dataset in Roweisfaces are shown in Fig. \ref{figure_reconstructed}. 
The quality of reconstruction falls down as $r_2$ is increased. We explain the reason in the following. 
According to Eq. (\ref{equation_reconstruction_severalPoint}), the reconstruction error for $\b{X}\b{A}$ is $||\b{X}\b{A} - \b{U}\,\b{U}^\top\b{X}\b{A}||_F^2$ where $\b{A}$ is a symmetric matrix. 
Minimizing this error where the bases are orthonormal is:
\begin{equation}\label{equation_optimization_reconstructionError}
\begin{aligned}
& \underset{\b{U}}{\text{minimize}}
& & ||\b{X}\b{A} - \b{U}\,\b{U}^\top\b{X}\b{A}||_F^2, \\
& \text{subject to}
& & \b{U}^\top \b{U} = \b{I},
\end{aligned}
\end{equation}
whose Lagrangian is simplified to $\mathcal{L} = \textbf{tr}(\b{A}^2 \b{X}^\top \b{X} - \b{X} \b{A}^2 \b{X}^\top \b{U} \b{U}^\top)$ noticing the constraint. 
Setting the derivative of Lagrangian to zero results in $\b{X} \b{A}^2 \b{X}^\top \b{U} = \b{U} \b{\Lambda}$ which is the eigenvalue problem for $\b{\b{X} \b{A}^2 \b{X}^\top}$. Comparing this to the solution of Eq. (\ref{equation_optimization_RDA}) and noticing Eq. (\ref{equation_R1}), shows we can have $\b{A}^2 = \b{H} \b{P} \b{H}$, $r_2=0$, and $\b{R}_2=\b{I}$ for minimization of reconstruction error. Hence, the best setting for reconstruction is to have $r_2=0$. 
In addition, if $r_1=0$, the objective in Eq. (\ref{equation_optimization_reconstructionError}) becomes the error between $\breve{\b{X}}$ and $\widehat{\b{X}}=\b{U}\b{U}^\top \breve{\b{X}}$ (see Eq. (\ref{equation_reconstruction_severalPoint})) which is the reconstruction error of centered data. 
This explains why PCA (with $r_2=0$) is the best linear method for reconstruction.

\subsection{RDA for Classification}

\begin{figure}[!t]
\centering
\includegraphics[width=3in]{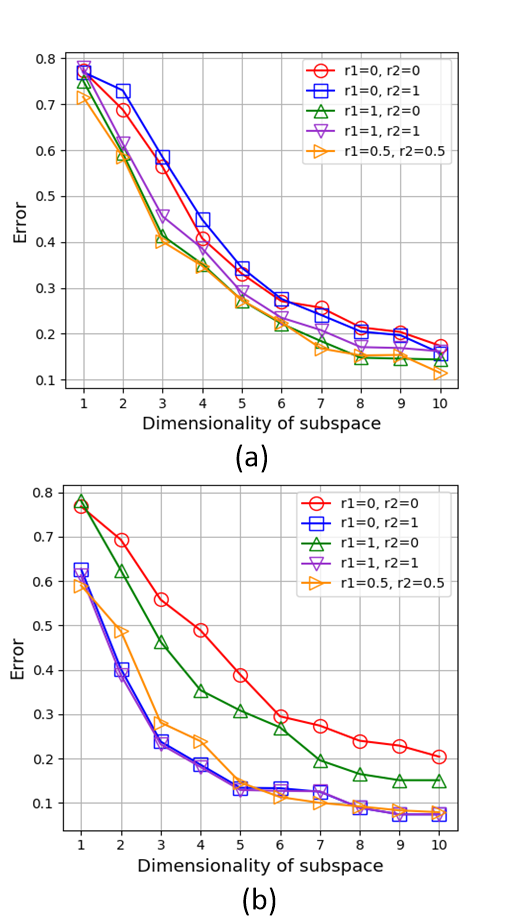}
\caption{The error rates for the $1$-NN classification of the subset of MNIST dataset: (a) some cases in RDA, (b) some cases in kernel RDA.}
\label{figure_MNIST_error_rates}
\end{figure}

To experiment on a classification task, we used a subset of the MNIST handwritten digit dataset \cite{web_MNIST_dataset} with $5000$ training and $1000$ test images. 
Note that since RDA uses a type of subspace learning based on eigenvalue and generalized eigenvalue problems, it cannot handle very large datasets. 
Prior works which make use of eigenvalue decomposition, such as \cite{barshan2011supervised,abdi2019discriminant}, have similar constraints.
The data were standardized. We used $1$-Nearest Neighbor ($1$-NN) classifier for the projected data instances because it shows the structure of embedded data. The error rates of the classification for some special cases of RDA and kernel RDA (with RBF kernel for $\b{K}_x$) are shown in Fig. \ref{figure_MNIST_error_rates} where we sweep over the dimensionality of subspace. 
As expected, in most cases, the kernel RDA performed better than RDA. For RDA, the DSDA and the case $r_1=r_2=0.5$ were better than PCA and FDA which makes sense because DSDA has full supervision level. Also, kernel DSDA and the kernel method for $r_1=r_2=0.5$ performed much better than kernel PCA and kernel SPCA. 

\subsection{RDA for Regression}

According to \cite{li2005contour} and inspired by \cite{barshan2011supervised}, we used three synthetic benchmarks which are for evaluating regression tasks in subspace learning. Let $\b{X}_{:i}$ and $\b{X}_{j:}$ denote the $i$-th column (instance) and $j$-th row (feature) of $\b{X}$, respectively. 
In the benchmark 1, $\b{X} \in \mathbb{R}^{4 \times n}$ and the vector of labels is $\b{y} = \b{X}_{1:} / (0.5 + (\b{X}_{2:} + 1.5)^2) + (1 + \b{X}_{2:})^2 + 0.5 \varepsilon$ where $\varepsilon \sim \mathcal{N}(0,1)$ is the Gaussian additive noise and $\b{X}_{:i} \sim \mathcal{N}(\b{0}, \b{I})$. 
In benchmark 2, $\b{X} \in \mathbb{R}^{4 \times n}$ and $\b{y} = \sin^2(\pi \b{X}_{2:} + 1) + 0.5\varepsilon$ where $\b{X}_{:i}$ is uniformly distributed on the set $[0,1]^4 \backslash \{\b{x} \in \mathbb{R}^4 | \b{x}_{j:} \leq 0.7, \forall j \in \{1,2,3,4\}\}$. the distribution of this benchmark is not elliptical. 
In benchmark 3, $\b{X} \in \mathbb{R}^{10 \times n}$ and $\b{y} = 0.5(\b{X}_{1:})^2 \varepsilon$ where the noise is multiplicative here and $\b{X}_{:i} \sim \mathcal{N}(\b{0}, \b{I})$. 
Inspired by \cite{barshan2011supervised}, we drew $50$ samples of size $100$ from these benchmark distributions from each of which we took $70\%$ and $30\%$ of data for training and test, respectively. We used linear regression and the top two features of the projected data. 
We used RBF kernel for both $\b{K}_x$ and $\b{K}_y$ (note that the delta kernel was useful for classification task). 

The average root mean squared errors of regression over the $50$ created datasets are reported in Table \ref{table_RMSE_regression} where only the cases with $r_2=0$ are used in RDA and kernel RDA because as explained before, regression only applies to $r_1=0$ in the Roweis map. 
As expected, often, the greater $r_1$ had better performance for the sake of larger supervision level. 
Moreover, in most cases, kernel RDA performs better than or similar to RDA.

\begin{table}[!t]
\caption{The average root mean squared error for regression experiments of RDA and Kernel RDA (KRDA) on the benchmark datasets.}
\label{table_RMSE_regression}
\renewcommand{\arraystretch}{1.3}  
\centering
\scalebox{0.85}{    
\begin{tabular}{l || l | c | c | c}
\hline
\hline
& & benchmark 1 & benchmark 2 & benchmark 3  \\
\hline
\hline
\multirow{3}{*}{RDA} & $r_1=0, r_2=0$ & 2.004 $\pm$ 0.673 & 0.155 $\pm$ 0.039 & 0.526 $\pm$ 0.413 \\
& $r_1=0.5, r_2=0$ & 1.556 $\pm$ 0.446 & 0.055 $\pm$ 0.021 & 0.558 $\pm$ 0.443 \\
& $r_1=1, r_2=0$ & 1.538 $\pm$ 0.441 & 0.048 $\pm$ 0.014 & 0.567 $\pm$ 0.452 \\
\hline
\multirow{3}{*}{KRDA} & $r_1=0, r_2=0$ & 2.061 $\pm$ 0.701 & 0.155 $\pm$ 0.039 & 0.521 $\pm$ 0.413 \\
& $r_1=0.5, r_2=0$ & 1.630 $\pm$ 0.632 & 0.054 $\pm$ 0.021 & 0.503 $\pm$ 0.394 \\
& $r_1=1, r_2=0$ & 1.615 $\pm$ 0.632 & 0.048 $\pm$ 0.014 & 0.493 $\pm$ 0.390 \\
\hline
\hline
\end{tabular}%
}
\end{table}

\section{Summary and Conclusion}\label{section_conclusion}

In this paper, we proposed RDA which generalized subspace learning including PCA, SPCA, and FDA. One of the extreme special cases is not yet in the literature and we named it Double Supervised Discriminant Analysis (DSDA). Dual RDA for some special cases was proposed here, too. Kernel RDA was also proposed using two methods of kernel trick and representation theory. Applying RDA and kernel RDA on facial dataset gave us Roweisfaces and kernel Roweisfaces generalizing eigenfaces, Fisherfaces, kernel eigenfaces, kernel Fisherfaces, and supervised eigenfaces.
We also demonstrated some cases where DSDA, the novel method derived from the bottom-right corner of the Roweis map, provides superior preliminary results.

Our analysis of the theory of RDA and kernel RDA shed light to some facts about some of the existing subspace learning methods:
(1) We showed that if $r_2=0$, the labels can be for either classification or regression. This explains why SPCA (with $r_2=0$) can be used for both classification and regression while FDA (with $r_2=1$) is only for classification task. 
(2) We showed that if $r_2=0$, the method has a dual which explains why PCA and SPCA (with $r_2=0$) have their dual methods but FDA (with $r_2=1$) does not have a dual. 
(3) If $r_2=0$ we have the dual required for the kernel trick. We also showed that only for $r_1=0$ and $r_1=1$, the inner product of the data points apear in the dual method. This explains why PCA (with $r_1=r_2=0$) and SPCA (with $r_1=1, r_2=0$) have their kernel methods using kernel trick but kernel FDA (with $r_1=0, r_2=1$) uses representation theory but not kernel trick.
(4) If $r_2=0$, we showed that the reconstruction error can be minimized. This explains why PCA (with $r_2=0$) is the best linear method for reconstruction error.

\appendices

\section{Derivation of Eq. (\ref{equation_d_W_pulled})}\label{section_appendix_derivation_N}

The mean of the $j$-th class, Eq. (\ref{equation_mean_of_class}), in the feature space is:
\begin{align}\label{equation_mean_of_class_inFeatureSpace}
\mathbb{R}^{t} \ni \b{\phi}(\b{\mu}_j) := \frac{1}{n_j} \sum_{i=1}^{n_j} \b{\phi}(\b{x}_i^{(j)}).
\end{align}
We have:
\begin{align*}
&\b{\phi}(\b{u})^\top \b{\Phi}(\b{S}_W)\, \b{\phi}(\b{u}) \\
&\overset{(a)}{=} \Big( \sum_{\ell=1}^n \theta_\ell\, \b{\phi}(\b{x}_\ell)^\top \Big) \Big( \sum_{j=1}^c \sum_{i=1}^{n_j} \big(\b{\phi}(\b{x}_i^{(j)}) - \b{\phi}(\b{\mu}_j)\big) \nonumber \\
&~~~~~~~~~~~~~~~~~~~ \big(\b{\phi}(\b{x}_i^{(j)}) - \b{\phi}(\b{\mu}_j)\big)^\top \Big) \Big( \sum_{k=1}^n \theta_k\, \b{\phi}(\b{x}_k) \Big) \\
&= \sum_{j=1}^c \sum_{\ell=1}^n \sum_{i=1}^{n_j} \sum_{k=1}^n \Big( \theta_\ell\, \b{\phi}(\b{x}_\ell)^\top \big(\b{\phi}(\b{x}_i^{(j)}) - \b{\phi}(\b{\mu}_j)\big) \\
&~~~~~~~~~~~~~~~~~~~ \big(\b{\phi}(\b{x}_i^{(j)}) - \b{\phi}(\b{\mu}_j)\big)^\top \theta_k\, \b{\phi}(\b{x}_k) \Big) 
\end{align*}

\begin{align*}
&\overset{(\ref{equation_mean_of_class_inFeatureSpace})}{=} \sum_{j=1}^c \sum_{\ell=1}^n \sum_{i=1}^{n_j} \sum_{k=1}^n \\
&~~~~~~~~~~~~~~~~~~~ \Big( \theta_\ell\, \b{\phi}(\b{x}_\ell)^\top \big(\b{\phi}(\b{x}_i^{(j)}) - \frac{1}{n_j} \sum_{e=1}^{n_j} \b{\phi}(\b{x}_e^{(j)})\big) \\
&~~~~~~~~~~~~~~~~~~~ \big(\b{\phi}(\b{x}_i^{(j)}) - \frac{1}{n_j} \sum_{z=1}^{n_j} \b{\phi}(\b{x}_z^{(j)})\big)^\top \theta_k\, \b{\phi}(\b{x}_k) \Big) 
\end{align*}

\begin{align*}
&\overset{(\ref{equation_kernel_scalar})}{=} \sum_{j=1}^c \sum_{\ell=1}^n \sum_{i=1}^{n_j} \sum_{k=1}^n \\
&~~~~~~~~~~~~~~~~~~~ \Big( \theta_\ell\, k(\b{x}_\ell, \b{x}_i^{(j)}) - \frac{1}{n_j} \sum_{e=1}^{n_j} \theta_\ell\, k(\b{x}_\ell, \b{x}_e^{(j)}) \Big) \\
&~~~~~~~~~~~~~~~~~~~ \Big( \theta_k\, k(\b{x}_i^{(j)}, \b{x}_k) - \frac{1}{n_j} \sum_{z=1}^{n_j} \theta_k\, k(\b{x}_z^{(j)}, \b{x}_k) \Big) 
\end{align*}

\begin{align*}
&\overset{(b)}{=} \sum_{j=1}^c \sum_{\ell=1}^n \sum_{i=1}^{n_j} \sum_{k=1}^n \\
&~~~~~~~~~~~~~~~~~~~ \Big( \theta_\ell\, k(\b{x}_\ell, \b{x}_i^{(j)}) - \frac{1}{n_j} \sum_{e=1}^{n_j} \theta_\ell\, k(\b{x}_\ell, \b{x}_e^{(j)}) \Big) \\
&~~~~~~~~~~~~~~~~~~~ \Big( \theta_k\, k(\b{x}_k, \b{x}_i^{(j)}) - \frac{1}{n_j} \sum_{z=1}^{n_j} \theta_k\, k(\b{x}_k, \b{x}_z^{(j)}) \Big) 
\end{align*}

\begin{align*}
&= \sum_{j=1}^c \sum_{\ell=1}^n \sum_{i=1}^{n_j} \sum_{k=1}^n \\
&\Big( \theta_\ell\, \theta_k\, k(\b{x}_\ell, \b{x}_i^{(j)})\, k(\b{x}_k, \b{x}_i^{(j)}) \\
&- \frac{2\,\theta_\ell\, \theta_k}{n_j} \sum_{z=1}^{n_j} k(\b{x}_\ell, \b{x}_i^{(j)})\, k(\b{x}_k, \b{x}_z^{(j)}) \\
& + \frac{\theta_\ell\, \theta_k}{n_j^2} \sum_{e=1}^{n_j} \sum_{z=1}^{n_j} k(\b{x}_\ell, \b{x}_e^{(j)})\, k(\b{x}_k, \b{x}_z^{(j)}) \Big)
\end{align*}

\begin{align*}
&= \sum_{j=1}^c \sum_{\ell=1}^n \sum_{i=1}^{n_j} \sum_{k=1}^n \\
&\Big( \theta_\ell\, \theta_k\, k(\b{x}_\ell, \b{x}_i^{(j)})\, k(\b{x}_k, \b{x}_i^{(j)}) \\
&- \frac{\theta_\ell\, \theta_k}{n_j} \sum_{z=1}^{n_j} k(\b{x}_\ell, \b{x}_i^{(j)})\, k(\b{x}_k, \b{x}_z^{(j)}) \Big) 
\end{align*}

\begin{align*}
&= \sum_{j=1}^c \bigg( \sum_{\ell=1}^n \sum_{i=1}^{n_j} \sum_{k=1}^n \Big( \theta_\ell\, \theta_k\, k(\b{x}_\ell, \b{x}_i^{(j)})\, k(\b{x}_k, \b{x}_i^{(j)}) \Big) \\
&- \sum_{\ell=1}^n \sum_{i=1}^{n_j} \sum_{k=1}^n \Big( \frac{\theta_\ell\, \theta_k}{n_j} \sum_{z=1}^{n_j} k(\b{x}_\ell, \b{x}_i^{(j)})\, k(\b{x}_k, \b{x}_z^{(j)}) \Big) \bigg) 
\end{align*}

\begin{align*}
&\overset{(\ref{equation_K_j})}{=} \sum_{j=1}^c \big( \b{\theta}^\top \b{K}_j \b{K}_j^\top \b{\theta} - \b{\theta}^\top \b{K}_j \frac{1}{n_j} \b{1}\b{1}^\top \b{K}_j^\top \b{\theta} \big) \\
&= \sum_{j=1}^c \b{\theta}^\top \b{K}_j \big( \b{I} - \frac{1}{n_j} \b{1}\b{1}^\top \big) \b{K}_j^\top \b{\theta} \\
&\overset{(\ref{equation_H_j})}{=} \sum_{j=1}^c \b{\theta}^\top \b{K}_j \b{H}_j \b{K}_j^\top \b{\theta} = \b{\theta}^\top \Big( \sum_{j=1}^c \b{K}_j \b{H}_j \b{K}_j^\top \Big) \b{\theta} \\
&\overset{(\ref{equation_N})}{=} \b{\theta}^\top \b{N} \b{\theta},
\end{align*}
where $(a)$ is because of Eqs. (\ref{equation_within_scatter}) and (\ref{equation_phi_u}) and $(b)$ is because $k(\b{x}_1, \b{x}_2) = k(\b{x}_2, \b{x}_1) \in \mathbb{R}$. $~~~$ Q.E.D.

\ifCLASSOPTIONcaptionsoff
  \newpage
\fi



%



\bibliographystyle{IEEEtran}
\bibliography{references}

%

\begin{IEEEbiography}[{\includegraphics[width=1in,height=1.25in,clip,keepaspectratio]{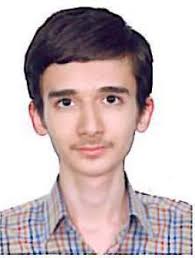}}]%
{Benyamin Ghojogh}
received the first and second B.Sc. degrees in electrical engineering (electronics and telecommunications) from the Amirkabir University of Technology, Tehran, Iran, in 2015 and 2017, respectively, and the M.Sc. degree in electrical engineering from the Sharif University of Technology, Tehran, Iran, in 2017. He is currently pursuing the Ph.D. degree in electrical and computer engineering with the University of Waterloo, Waterloo, ON, Canada. His research interests include machine learning, manifold learning, and statistical learning.
\end{IEEEbiography}

\begin{IEEEbiography}[{\includegraphics[width=1in,height=1.25in,clip,keepaspectratio]{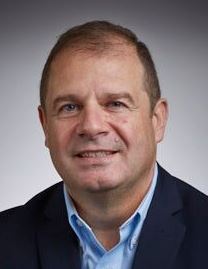}}]%
{Fakhri Karray}
is UW Research Chair Professor in Electrical and Computer Engineering and co-Director of the Pattern Analysis and Machine Intelligence Center. He received his Ph.D. from the University of Illinois, USA (1989), in the area of systems and control. Dr. Karray’s research interests are in the areas of intelligent mechatronics and transportation systems, soft computing, autonomous machines and natural man–machine interaction. He is the author of more than 300 technical articles, 14 US patents, a major textbook on soft computing and more than 20 textbook chapters. He has chaired/co-chaired more than 15 international conferences. He has also served as the associate editor/guest editor for more than 10 journals, including the IEEE Transactions on Systems Man Cybernetics (B), the IEEE Transactions on Neural Networks and Learning Systems, the IEEE Transactions on Mechatronics, and the IEEE Computational Intelligence Magazine. He is the Chair of the IEEE Computational Intelligence Society Chapter in Waterloo, Canada.
\end{IEEEbiography}

\begin{IEEEbiography}[{\includegraphics[width=1in,height=1.25in,clip,keepaspectratio]{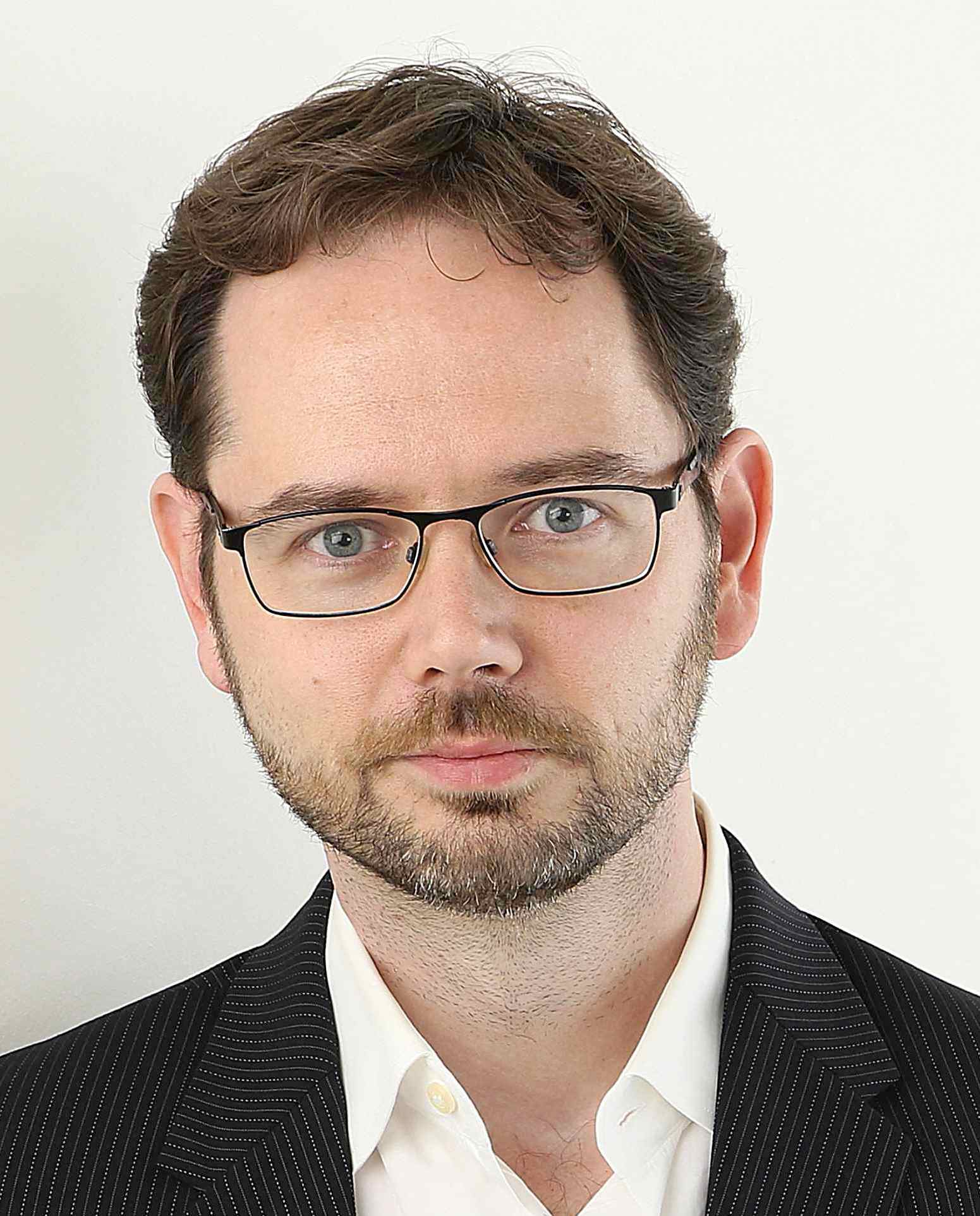}}]%
{Mark Crowley}
received the BA degree in computer science from York University, Toronto, Canada, the MSc degree from the University of British Columbia, and the PhD degree in computer science from the University of British Columbia, Vancouver, Canada, in 2011. His research interests include machine learning, reinforcement learning, deep learning, ensemble methods, and computational sustainability. He is currently an assistant professor at the University of Waterloo, Waterloo, Canada.
\end{IEEEbiography}




\end{document}